%% file: iclr2027_conference.tex
\documentclass{article}
\usepackage[preprint]{iclr2027_conference}
\usepackage{times}

\usepackage[utf8]{inputenc} % allow utf-8 input
\usepackage[T1]{fontenc}    % use 8-bit T1 fonts
\usepackage{url}            % simple URL typesetting
\usepackage{booktabs}       % professional-quality tables
\usepackage{amsfonts}       % blackboard math symbols
\usepackage{microtype}      % microtypography
\usepackage{xcolor}         % colors

\usepackage{graphicx}
\usepackage{amsmath}
\usepackage{amssymb}
\usepackage{mathtools}
\usepackage{multirow}
\usepackage{float}
\usepackage{wrapfig}

\input{custom_commands.tex}
\usepackage{hyperref}
\hypersetup{colorlinks=true, allcolors=blue}

\title{VLM-Guided Experience Replay}

\author{%
  Elad Sharony\thanks{Corresponding author: \href{mailto:eladsharony@campus.technion.ac.il}{\nolinkurl{eladsharony@campus.technion.ac.il}}.} \\
  Technion \\
  \And
  Tom Jurgenson \\
  Technion \\
  \And
  Orr Krupnik \\
  Technion \\
  \AND
  Dotan Di Castro \\
  ForSight Robotics \\
  \And
  Shie Mannor \\
  Technion, Nvidia Research \\
}

\begin{document}

\maketitle

\begin{abstract}
Recent advances in Large Language Models (LLMs) and Vision-Language Models (VLMs) have enabled powerful semantic and multimodal reasoning capabilities, creating new opportunities to enhance sample efficiency, high-level planning, and interpretability in reinforcement learning (RL).
While prior work has integrated LLMs and VLMs into various components of RL, the replay buffer, a core component for storing and reusing experiences, remains unexplored.
We propose addressing this gap by leveraging VLMs to guide the prioritization of experiences in the replay buffer.
Our key idea is to use a frozen, pre-trained VLM as an automated evaluator to identify and prioritize promising sub-trajectories from the agent's experiences.
Across scenarios, including game-playing and robotics, spanning both discrete and continuous domains, agents trained with our proposed prioritization method achieve \textbf{15--57\%} higher average success rates and improve sample efficiency by \textbf{35--55\%} compared to previous approaches (Fig.~\ref{fig:budget-frontier}).
\end{abstract}
\ificlrfinal
\vspace{-4pt}
{\centering Project page: \href{https://esharony.me/projects/vlm-rb/}{esharony.me/projects/vlm-rb}\par}
\fi

\begin{figure}[h]
    \centering
    \includegraphics[width=0.7\linewidth,trim=-20pt 0 20pt 0]{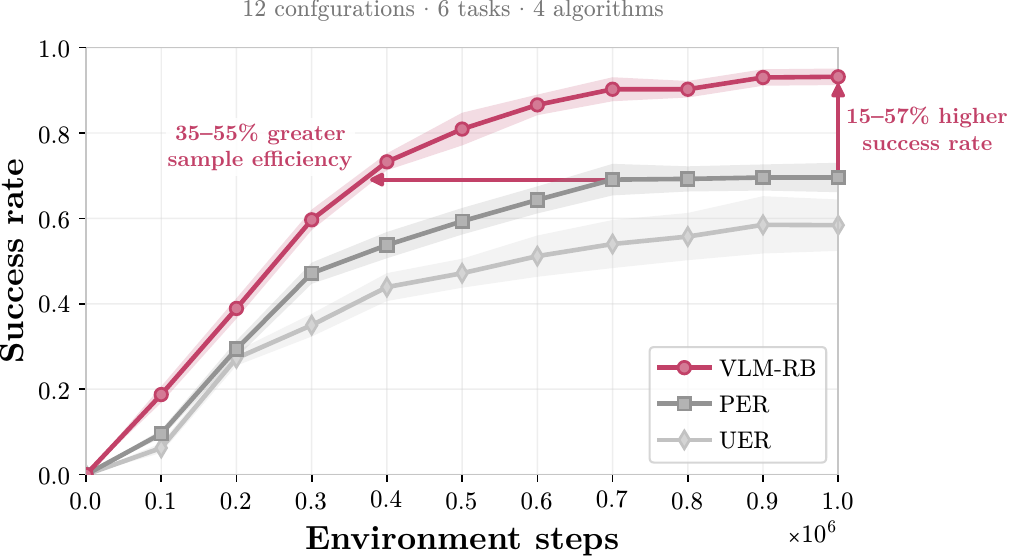}
    \caption{\textbf{\fmethod learns more from the same interaction budget.}
    Mean success across 12 algorithm--task configurations, averaged across 5 seeds. Shading denotes standard error.}
    \label{fig:budget-frontier}
\end{figure}

\section{Introduction}
\label{sec:intro}

\looseness=-2
Reinforcement learning (RL) has achieved impressive results across domains ranging from robotics to natural language systems~\citep{levine2016end, ouyang2022training} and logistics management~\citep{tang2025deep}.
Despite this success, RL remains notoriously sample-inefficient, sensitive to hyperparameters, and heavily dependent on carefully designed reward functions~\citep{kalashnikov2021scaling, kroemer2021review, zhu2020ingredients}.

A well-established line of research—off-policy RL~\citep{lin1992self, mnih2013playing}—addresses the efficiency bottleneck through experience reuse.
During training, transitions are collected into a dataset known as a \emph{replay buffer}, and sampled multiple times to update the policy.
Ideally, a sampling mechanism would prioritize the most meaningful transitions: those that provide the richest learning signal, irrespective of when they were collected. The core challenge is to identify which experiences are truly meaningful for the learning process.

\looseness=-2
Conventional methods such as Prioritized Experience Replay (\fpert, \citealt{schaul2015prioritized}) approximate this ideal by prioritizing transitions with high temporal-difference (TD) error. The TD error measures the discrepancy between a bootstrapped value estimate and its target~\citep{sutton1998reinforcement}. In effect, this approach promotes transitions which are misaligned with the agent's future outcome predictions.
While TD error prioritization can correct value estimates, it is fundamentally limited by its lack of semantic awareness. It cannot distinguish transitions which reflect genuine progress toward task completion from those that do not.

\looseness=-2
This limitation is particularly evident in long-horizon, sparse-reward tasks, such as robotic manipulation in OGBench~\citep{park2024ogbench}.
For instance, critical transitions—such as unlatching a door or grasping a tool—may yield low TD errors early in training. This is a consequence of delayed credit assignment preventing the reward signal from propagating to these earlier states.
Conversely, visually distinct but task-irrelevant motions can yield large TD errors, despite contributing little to actual task completion.

\begin{figure}[h]
    \centering
    \includegraphics[width=0.9\linewidth]{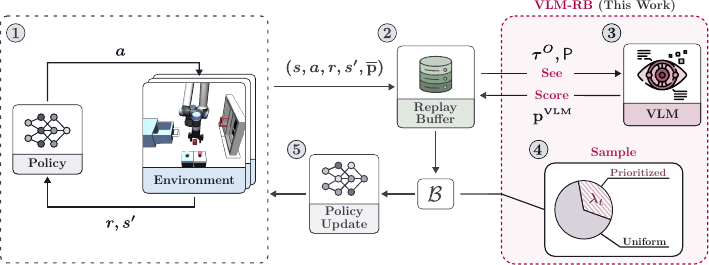}
    \caption{\looseness=-2
    \textbf{System diagram:} (1) Data is collected with the current policy $\pi_k$ interacting with multiple instances of the environment. (2) Transitions $(s,a,r,s')$ are kept in a prioritized replay buffer with a provisional VLM score $\bar{\bm{\mathrm{p}}}$. (3) Asynchronously, after each insertion, a VLM worker scores the corresponding rendered clip $\tau^O$ under a prompt $\mathsf{P}$ and writes the resulting score $\bm{\mathrm{p}}^{\scriptscriptstyle\mathrm{VLM}}$ back to the replay buffer (see Section~\ref{sub-sec:mechanism}). (4) The learner samples a minibatch using the mixture distribution $q_t$ which interpolates between uniform and VLM-prioritized replay (see Section~\ref{sub-sec:mechanism}). (5) The agent policy is updated to $\pi_{k+1}$.}
    \label{fig:vlmrb-method}
\end{figure}

To bridge this gap, we propose replacing heuristic inductive biases with external sources of semantic understanding.
\looseness=-2
Recent literature highlights the potential of integrating large-scale Vision-Language Models (VLMs) with RL to leverage pre-trained world knowledge~\citep{liang2024survey, zhang2024vision}.
Because VLMs combine visual perception with language reasoning, they can interpret complex environments and mitigate RL-specific challenges—such as sparse rewards or inefficient exploration—by providing enriched context~\citep{cao2024survey, schoepp2025evolving}. This motivates our primary research question:
\begin{tcolorbox}[
    colback=gray!5!white,
    colframe=gray!75!black,
    boxrule=0.5pt,
    arc=2pt,
    left=4pt, right=4pt, top=5pt, bottom=5pt,
    boxsep=0pt
]
    \centering
    \textit{Can the semantic knowledge of pre-trained VLMs be used to prioritize meaningful experiences in replay buffers?}
\end{tcolorbox}

To answer this, we introduce \fmethod (Fig.\ref{fig:vlmrb-method}), which integrates a pre-trained VLM directly into the experience prioritization pipeline.
In contrast to heuristics relying on statistical proxies such as TD error, uncertainty \mbox{\citep{carrasco2025uncertainty}}, or density ratios~\citep{zhao2019curiosity}, \fmethod leverages the reasoning capabilities of VLMs to \emph{score} experiences according to their semantic relevance.

We design this framework as a modular \textit{plug-and-play} prioritization layer that can be integrated with any off-policy algorithm that uses a replay buffer.
By using frozen off-the-shelf VLMs, we leverage strong semantic priors while avoiding the computational cost of fine-tuning or additional inference latency at deployment.
Within the training pipeline, \fmethod promotes efficiency by querying the VLM asynchronously, thereby sidestepping the blocking inference bottleneck that often constrains the throughput of LLM/VLM-enhanced RL methods.

\looseness=-2
We empirically show that pre-trained VLMs yield semantically grounded prioritization signals which are closely aligned with task objectives. Leveraging this semantic guidance, \fmethod consistently outperforms existing baselines in both discrete game-playing and simulated continuous-control manipulation. Importantly, the moderate throughput overhead (about 12\%) is more than offset by substantial gains in sample efficiency (35--55\%) and average success rates (15--57\%).

\section{Preliminaries}\label{sec:prelim}

\textbf{Off-Policy Reinforcement Learning.}
\looseness=-2
We consider the standard RL framework, where an agent interacts with an environment modeled as an infinite-horizon Markov Decision Process (MDP)
$\mathcal{M} = (\mathcal{S}, \mathcal{A}, P, r, \gamma, \rho_0)$,
with state space $\mathcal{S}$, action space $\mathcal{A}$ (either discrete or continuous), transition kernel $P: \mathcal{S} \times \mathcal{A} \rightarrow \Delta(\mathcal{S})$, reward function $r: \mathcal{S} \times \mathcal{A} \rightarrow \mathbb{R}$, discount factor $\gamma \in (0,1)$ and initial-state distribution $\rho_0 \in \Delta(\mathcal{S})$.
We further assume access to a rendering function $\psi: \mathcal{S} \to \mathcal{O} \subseteq \mathbb{R}^{H \times W \times 3}$ which maps states to $H\times W$-sized visual observations $o_t = \psi(s_t)$.
The agent learns a stochastic policy $\pi_\theta: \mathcal{S} \rightarrow \Delta(\mathcal{A})$ to maximize the expected discounted return
\(
J(\pi) = \mathbb{E}_{\tau \sim p_\pi(\tau)}\left[\sum_{t=0}^\infty \gamma^t r_t \right],
\)
where $\tau = (s_0, a_0, r_0, s_1, a_1, r_1, \ldots)$ denotes a trajectory and $p_\pi(\tau)$ is the trajectory distribution induced by policy $\pi$.

\looseness=-2
To evaluate and improve policies, value functions are used to estimate expected returns.
The \emph{action-value function} (Q-function) $Q^{\pi}: \mathcal{S} \times \mathcal{A} \rightarrow \mathbb{R}$ represents the expected discounted return from state $s$ after taking action $a$ and then following policy $\pi$:
\(
Q^{\pi}(s,a) = \mathbb{E}_{\tau \sim p_\pi(\tau | s_0 = s, a_0 = a)}\left[\sum_{t=0}^\infty \gamma^t r_t \right].
\)
In modern RL methods, $Q^{\pi}$ is typically a learned function (represented by a neural network), denoted the \emph{critic}.

In off-policy RL, rather than exclusively using on-policy data, optimization is performed using previously collected experience stored in a \textit{replay buffer}~\citep{mnih2013playing}.
Let $\mathcal{D}_t$ denote the replay buffer at time $t$.
We sample training tuples $(s,a,r,s') \in \mathcal{D}_t$ according to a time-dependent distribution $q_t \in \Delta(\mathcal{D}_t)$.
In standard \emph{uniform} replay (\fuer), $q_t(i) = 1/|\mathcal{D}_t|$ for all $i \in \mathcal{D}_t$.
This formulation highlights that the optimization dynamics depend on the evolving sampling distribution $q_t$.

\textbf{Experience Replay Prioritization.}
As opposed to uniform replay, prioritized replay methods bias retrieval toward transitions expected to yield a stronger learning signal.
For clarity, we formalize this process as three distinct steps: \emph{scoring}, \emph{prioritization}, and \emph{sampling}.
\emph{Scoring} assigns each transition $i \in \mathcal{D}_t$ a scalar value $\bm{\mathrm{p}}_i \in \mathbb{R}$ reflecting its heuristic utility.
\emph{Prioritization} maps these raw scores into a probability distribution over the buffer, $q_t^{\mathrm{P}} \in \Delta(\mathcal{D}_t)$.
\emph{Sampling} then draws a minibatch of indices $\{i_k\}_{k=1}^B$ from a target distribution $q_t$, or a mixture of several such distributions.

\looseness=-2
Prioritized Experience Replay (\fpert, \citealt{schaul2015prioritized}) is a specific instance of this framework.
Its scoring function utilizes the temporal-difference (TD) error magnitude, $\bm{\mathrm{p}}_i = |\delta_i| + \varepsilon$ (where $\varepsilon>0$ ensures non-zero probability), where $\delta_i$ is the discrepancy between the value estimate and its target, namely $\delta_i=Q_\theta(s_t, a_t) - (r+\gamma Q_\theta(s_{t+1}, a'))$, where $\theta$ are the critic parameters and $a'$ is the action predicted by the current policy in $s_{t+1}$. In Q-learning methods, $a'$ is the argmax action, and in actor-critic methods for continuous control, it is $\pi(s_{t+1})$.
The prioritized distribution is defined as $q_t^{\mathrm{P}}(i)\propto \bm{\mathrm{p}}_i^\alpha$, where $\alpha\in[0,1]$ determines the degree of prioritization.
While \fpert{} effectively focuses updates on transitions with larger TD errors, it relies on value estimates which may be arbitrary and noisy early in training, particularly in sparse-reward settings.

\textbf{Vision--Language Models.}
\looseness=-2
Recent advances in large-scale representation learning have given rise to \emph{Vision--Language Models} (VLMs), which learn aligned embeddings of visual and textual modalities through joint pre-training on large-scale image--text corpora~\citep{radford2021learning, jia2021scaling, zhai2023sigmoid, li2023blip}.
These models have demonstrated strong generalization across a wide range of downstream tasks, including zero-shot recognition, goal specification, and reward inference~\citep{ahn2022can,rocamonde2023vision}.
While standard VLMs process static images, \emph{Video Question Answering} (Video-QA) models extend these capabilities to ingest a temporal sequence of frames (a video clip) along with a natural-language query, aggregating visual information across time to reason about events, motion, and causal relationships.
Concretely, for the rest of this paper, we employ Perception-LM~\citep{cho2025PerceptionLM}, a state-of-the-art open-source Video-QA family of models with multiple model sizes (1B/3B/8B).

\section{\fmethod}
\label{sec:method}

\looseness=-2
In this section, we present \fmethod (Fig.\ref{fig:vlmrb-method} and Algorithm~\ref{alg:vlmrb-async}), a plug-and-play framework designed for any off-policy algorithm that leverages a replay buffer. The central idea is to use a VLM to assign semantic scores to sub-trajectories, subsequently biasing the sampling distribution toward higher-scoring experiences during policy optimization.

\subsection{Scoring, Prioritization, and Sampling}
\label{sub-sec:mechanism}

We now define how semantic scores are extracted from collected data, how these scores induce a prioritization distribution, and how \fmethod samples from this distribution.

\looseness=-2
\textbf{Scoring.} \fmethod uses a pre-trained frozen VLM to extract semantic scores from collected experiences.
To enable this, we first construct temporal visual sequences from the agent's history.
Formally, let $\tau^O_i=(o_{i}, o_{i+1}, \ldots, o_{i+L-1})$ denote a visual clip comprising $L$ rendered frames.
The VLM maps this clip and a text prompt $\mathsf{P}$ to a scalar score:
\begin{equation}
\label{eq:vlm_scoring}
\bm{\mathrm{p}}_i^{\scriptscriptstyle\mathrm{VLM}}=f_{\scriptscriptstyle\mathrm{VLM}}(\tau_i^O, \mathsf{P})\in \mathbb{R},
\end{equation}
where different choices of $f_{\scriptscriptstyle\mathrm{VLM}}$ are detailed in \Appref{app:impl-details}.
The prompt $\mathsf{P}$ directs the VLM's scoring mechanism, leveraging the model's inherent world knowledge to identify meaningful behaviors even in the absence of dense reward signals. While this interface supports the injection of detailed and task-specific priors, we find that the VLM's intrinsic scene understanding is sufficient for our purposes. We therefore employ a general goal-agnostic, domain-generic prompt (see Appendix~\ref{app:prompts} for details).

\looseness=-2
This effectively labels useful sequences without the need for hand-crafted task-specific definitions.
Crucially, because the VLM is frozen and its evaluation depends only on static visual content, each clip is scored exactly once. This offers a significant efficiency advantage over methods such as \fpert, which require ongoing priority updates as the value function evolves.

\begin{wrapfigure}{r}{0.5\linewidth}
    \vspace{-12pt}
    \centering
    \includegraphics[width=\linewidth]{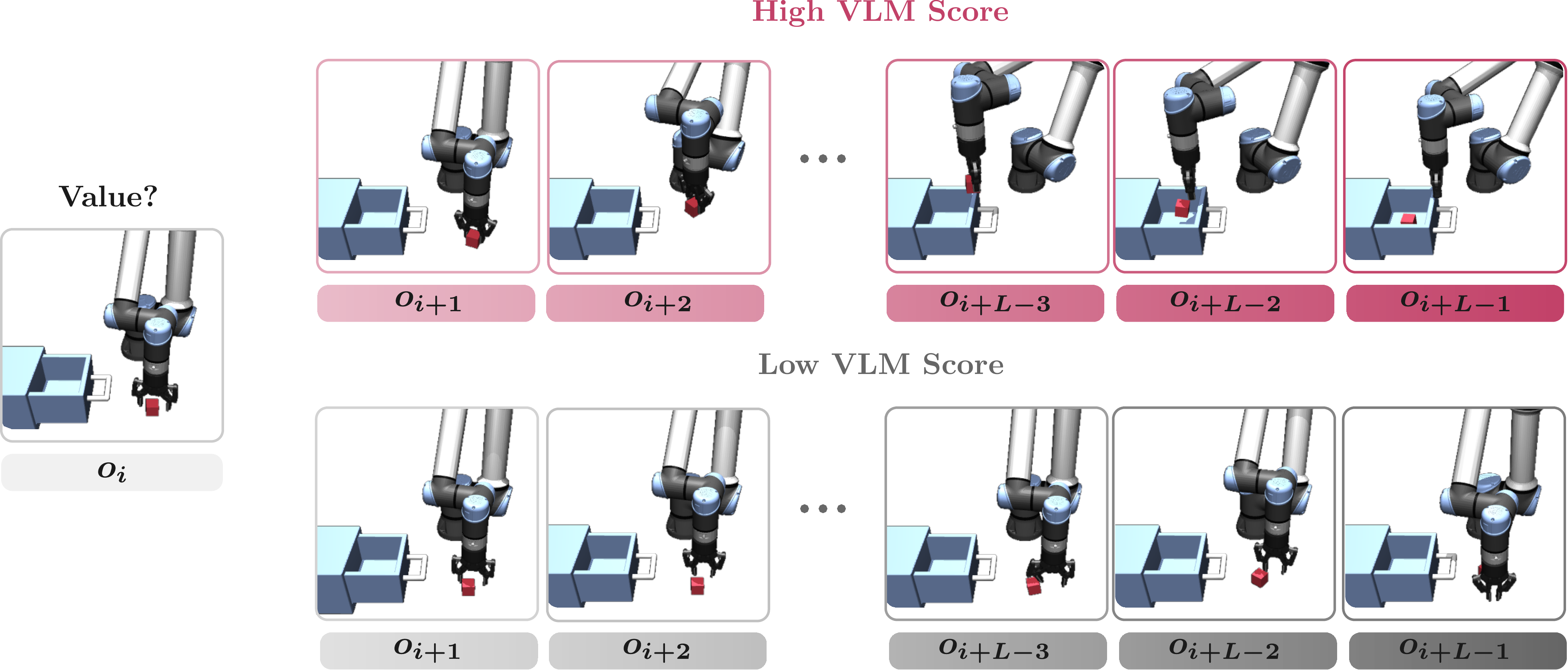}
    \caption{\textbf{Temporal context resolves visual ambiguity.} From the initial state $o_i$ (left), multiple futures are possible. The top sub-trajectory shows a successful grasp, while the bottom shows stagnation. By scoring sub-trajectories rather than single frames, the VLM has sufficient context to distinguish meaningful progress from failure modes.}
    \label{fig:frame_many_futures}
    \vspace{-8pt}
\end{wrapfigure}

A natural question is whether scoring individual frames, rather than clips, would suffice. We argue that single frames are fundamentally limited by \textit{semantic ambiguity}.
Consider Fig.~\ref{fig:frame_many_futures}: a single observation of a robotic gripper hovering above an object is visually identical across two vastly different temporal contexts, the onset of a successful grasp or the aftermath of a failed attempt.
\looseness=-2
Without temporal context a VLM cannot disambiguate these scenarios, potentially assigning high scores to frames which actually belong to failure modes.
By scoring sub-trajectories instead, we increase the likelihood of the VLM having sufficient temporal information to distinguish meaningful behaviors from failures.
While a sliding-window variant could potentially provide denser labels, we leave this (computationally intensive) alternative for future work.

\textbf{Prioritization.}
We propagate the VLM score of each clip to all transitions within that clip and combine it multiplicatively with the latest TD error. This defines our priority signal, which is converted into a distribution:
\begin{equation}
    \label{eq:vlm_td_priority}
    \begin{aligned}
    \bm{\mathrm{p}}_i &= \bm{\mathrm{p}}_i^{\scriptscriptstyle\mathrm{VLM}}\,|\delta_i|,
    \qquad
    q_t^{\mathrm{P}}(i)
    &= \frac{(\bm{\mathrm{p}}_i+\varepsilon)^\alpha}{\sum_{j\in\mathcal D_t}(\bm{\mathrm{p}}_j+\varepsilon)^\alpha
        }.
    \end{aligned}
\end{equation}
The two factors carry complementary information: $\bm{\mathrm{p}}^{\scriptscriptstyle\mathrm{VLM}}_{i}$ marks which experiences are semantically consistent with task success, whereas $|\delta_i|$ marks where the value function is still inaccurate.
\pagebreak

\textbf{Sampling.}
\looseness=-2
If we were to sample only from $q_t^{\mathrm{P}}$, updates would concentrate on the highest-scoring transitions. Because the VLM is an external prior and may score useful transitions inaccurately, we instead preserve full-buffer coverage by mixing VLM-guided prioritization with uniform replay:
\begin{equation}
    \label{eq:sampling_mixture}
    q_t(i) = \lambda_t q_t^{\mathrm{P}}(i) + \frac{1-\lambda_t}{|\mathcal D_t|},
\end{equation}
where $\lambda_t \in [0, 1]$ controls the strength of the VLM guidance.
In practice, each batch draws a $\lambda_t$ fraction from $q_t^{\mathrm{P}}$ and the remainder uniformly.
We use a linear warm-up schedule: starting with $\lambda_0=0$ (pure uniform sampling), we gradually anneal to $\lambda_{\max}=0.5$ over the first half of training.
The warm-up favors broad coverage early in training, while the uniform component limits the influence of inaccurate VLM scores throughout training. Removing this component is detrimental in our ablation (Section~\ref{sec:ablations}).

\subsection{Efficient Implementation}
\label{sub-sec:analysis}

\looseness=-2
Incorporating a VLM into the RL loop introduces significant computational overhead relative to standard components. We address this with two key design choices.
First, we leverage the decoupling between data collection and policy optimization in off-policy learning to asynchronously score experiences.
As illustrated in Fig.~\ref{fig:vlmrb-method}, the VLM interacts with the replay buffer in the background, ensuring that policy optimization is never blocked by inference latency.
Until a clip's VLM score becomes available, $\bm{\mathrm{p}}_i^{\scriptscriptstyle\mathrm{VLM}}$ is replaced by the running mean of previously returned VLM scores.
Second, we use a lightweight 1B parameter model~\citep{cho2025PerceptionLM} (see Appendix~\ref{app:vlm-size-ablation}).

\section{Experiments}
\label{sec:experiments}

\looseness=-2
In this section, we evaluate the efficacy of \fmethod in leveraging semantic priors for efficient exploration.
We first ask: does the VLM signal provide meaningful guidance for exploration? To answer this, we analyze its correlation with learned value estimates and its dependence on visual semantics (Section~\ref{sec:vlm_analysis}).
Next, we benchmark \fmethod against \fuer, \fpert, and alternative prioritization schemes, evaluating both performance and sample efficiency across a range of discrete and continuous control tasks (Section~\ref{sec:main_results}).
Finally, we examine which design choices most affect the performance of \fmethod, focusing on the prioritization signal, the sampling mixture, and VLM model size (Section~\ref{sec:ablations}).

\subsection{Experimental Setup}
\label{sec:experimental-setup}

\textbf{Tasks.}
\looseness=-2
We evaluate \fmethod on two domains with discrete and continuous action spaces: (i) \texttt{DoorKey} from \texttt{MiniGrid}~\citep{chevalier2023minigrid}, using grid sizes \texttt{8x8}, \texttt{12x12}, and \texttt{16x16} to vary exploration difficulty; and (ii) \texttt{scene} from OGBench~\citep{park2024ogbench}, using the predefined tasks $3$, $4$, and $5$, which require increasingly long-horizon compositional manipulation (unlocking/locking and coordinated object placement).
In all experiments, agents receive \emph{state-based} observations (rather than pixels) to minimize perception-related confounders and isolate the effect of replay prioritization on exploration and sample efficiency.
More details about the various tasks are provided in \Appref{app:environments}.

\textbf{Baselines.}
\looseness=-2
We compare \fmethod to two standard replay sampling methods: Uniform Experience Replay (\fuer), which samples transitions uniformly from the replay buffer, and \fpert, which prioritizes sampling by TD-error.
We report \fuer and \fpert comparisons for four different algorithms:
DQN~\citep{van2016deep}, IQN~\citep{dabney2018implicit}, TD3~\citep{fujimoto2018addressing}, and SAC~\citep{haarnoja2018soft}.
In addition, to compare \fmethod against alternative replay prioritization methods, we run a focused ablation for DQN comparing to Attentive Experience Replay (AER, ~\citealt{sun2020attentive}), Experience Replay Optimization (ERO, ~\citealt{zha2019experience}), and Reducible Loss Prioritization (ReLO, ~\citealt{sujit2023prioritizing}).
Implementation details for all baselines are provided in \Appref{app:impl-details}.

\textbf{Metrics.}
We report the \textsc{Success Rate} (SR), defined as the fraction of episodes in which the agent completes the task within the maximum episode length, \(T_{\mathrm{max}}\).
The \textsc{Average Success Rate} (ASR) is the mean SR over \(N=32\) evaluation episodes for a given seed.
Final results are the ASR averaged over \(M=5\) seeds, with the standard error of the mean (SEM) shown as shaded regions in plots.

\begin{figure}[H]
    \centering
    \includegraphics[width=\linewidth]{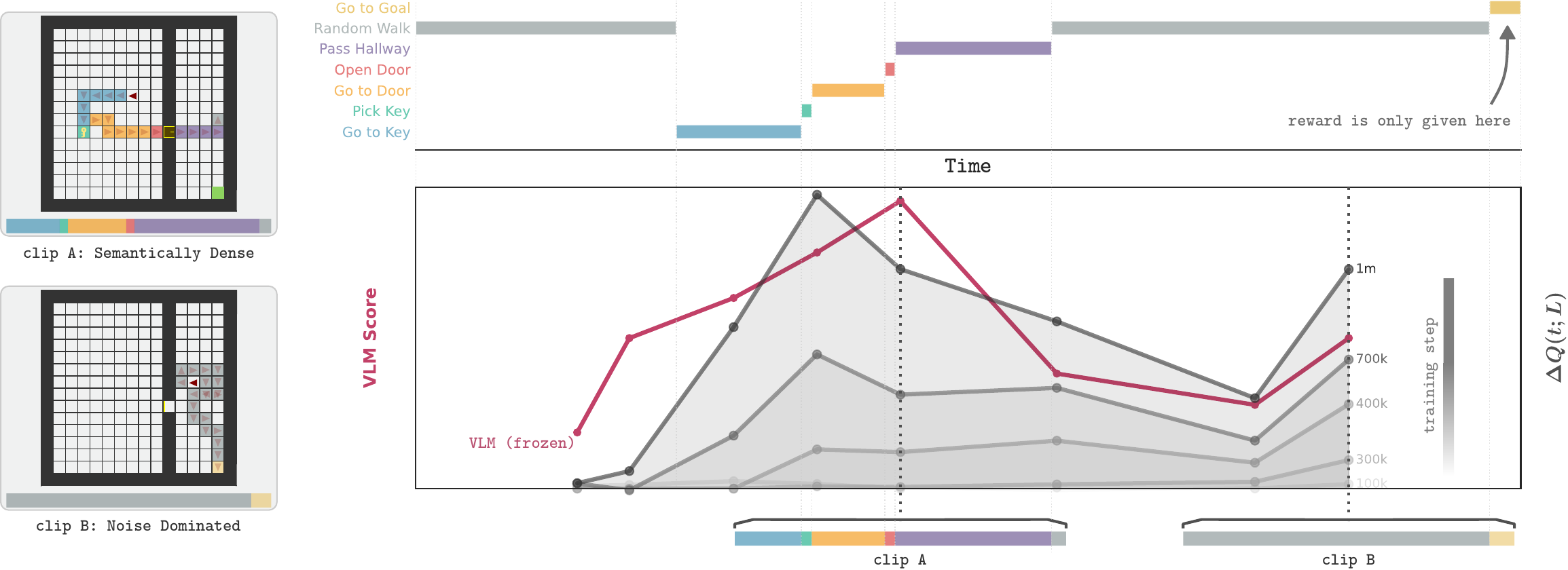}
    \caption{\textbf{Frozen VLM scoring qualitatively anticipates learned value.} We visualize a single reference episode in \texttt{MiniGrid/DoorKey-16x16} which contains both goal-directed actions and random noise. \textbf{Left:} Visualizations of two 32-frame sub-trajectories: \textbf{\texttt{clip A}} captures a semantically relevant sequence, while \textbf{\texttt{clip B}} shows a mostly random walk which eventually reaches the goal. \textbf{Top right:} A timeline of ground-truth events; sparse reward is only received at the final goal. Both right panels share the reference-episode timestep on the x-axis. \textbf{Bottom right:} The rose curve shows the frozen VLM score for $L=32$-frame clips. The gray curves show the temporal value difference, $\Delta Q(t; L) = Q(t+\tfrac{L}{2}) - Q(t-\tfrac{L}{2})$, calculated from critic checkpoints at increasing training steps (light to dark).}
    \label{fig:vlm_vs_Q}
    \vspace{-8pt}
\end{figure}

\subsection{Do VLMs Encode Useful Semantic Priors?}
\label{sec:vlm_analysis}

\begin{wrapfigure}{r}{0.5\linewidth}
    \vspace{-12pt}
    \centering
    \includegraphics[width=\linewidth]{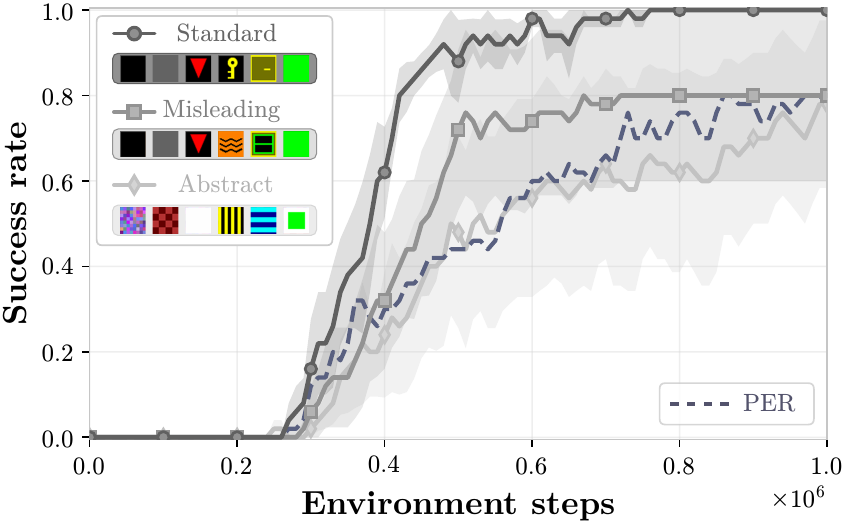}
    \caption{\looseness=-2
    \textbf{Success depends on alignment with semantic priors.} We compare the performance of \fmethod across frames rendered with \emph{Standard} visuals (dark circles), \emph{Misleading} swapped sprites (medium squares), and \emph{Abstract} textures (light diamonds). The dashed line indicates the ASR of \fpert. Crucially, the agent's actual observations and the underlying MDP are identical across all settings; only the visual input to the VLM is altered (see Appendix~\ref{sec:vlm_prior} for visual samples of these modifications). The significant performance drop in the \emph{Misleading} and \emph{Abstract} settings confirms that our method relies on the VLM correctly recognizing specific semantic objects (e.g., keys and doors) to accelerate learning.}
    \label{fig:vlm_prior_ablation}
    \vspace{-8pt}
\end{wrapfigure}

\looseness=-2
Before turning to downstream RL results, we first ask whether the frozen VLM offers a signal that is actually useful for prioritizing data.
To illustrate the semantic grounding of our prioritization, we compare the frozen VLM scores with critic value differences ($\Delta Q$) as learning progresses.
Concretely, we track how the value estimates of the critic evolve on a fixed reference episode, using checkpoints from a successful training run.
As shown in Fig.~\ref{fig:vlm_vs_Q}, the critic's estimates are initially flat and uninformative (light gray lines).
Over the course of training, the critic gradually learns to assign high value to the same semantic events (such as picking up a key or opening a door) that the VLM identified from the outset.
This observation suggests that \fmethod can immediately identify semantically relevant data, enabling faster learning than approaches which must wait for the critic to converge.

\looseness=-2
Having observed qualitative alignment between the VLM and a well-trained Q-value predictor on this reference episode, we next ask whether the signal reflects interpretable semantic priors.

\looseness=-2
To test whether \fmethod specifically leverages the VLM's pre-trained visual semantics, we employ the ``Modified Game'' paradigm~\citep{dubey2018investigating}.
We modify only the rendered frames used for VLM scoring, leaving the underlying MDP and agent observations untouched.
We consider three settings: (i) the \emph{Standard} game, with unaltered visuals; (ii) the \emph{Misleading} game, where key objects are swapped (for example, traps appear as goals); and (iii) the \emph{Abstract} game, where objects are replaced with random noise patterns (See Appendix~\ref{sec:vlm_prior} for visual examples).

As shown in Fig.~\ref{fig:vlm_prior_ablation}, the \emph{Misleading} and \emph{Abstract} variants significantly slow learning and increase variance.
Notably, in the \emph{Abstract} setting, performance degrades to the level of \fpert (dashed line).
Because the underlying MDP remains unchanged, this degradation demonstrates that the VLM's prioritization is only effective when the visual input aligns with natural semantic priors.

In summary, these results confirm that \fmethod scores are indicative of meaningful semantic behavior.
We now turn to the question: does this semantic guidance actually yield improved sample efficiency and asymptotic performance in off-policy RL?

\begin{table}[ht]
    % \vspace{-2pt}
    \centering
    \caption{
        Improvement with respect to baselines. \textbf{Performance} denotes the mean per-seed best ASR ($M=5$ seeds). \textbf{Sample efficiency} tracks the steps required to reach the \textit{baseline's} best performance; unreached targets are assigned the 1M-step horizon. Values in parentheses indicate the relative improvement over the baseline. Both metrics are averaged across the aggregated algorithms. The corresponding training curves are provided in Fig.~\ref{fig:training-curves}.}
    \label{tab:experiments-agg-algo}
    \footnotesize
    \setlength{\tabcolsep}{4pt}
    \resizebox{\linewidth}{!}{%
    \begin{tabular}{l c cc cc}
    \toprule
    & & \multicolumn{2}{c}{\textit{vs. \fuer}} & \multicolumn{2}{c}{\textit{vs. \fpert}} \\
    \cmidrule(lr){3-4}\cmidrule(lr){5-6}
    \textbf{Algorithm} & \textbf{Level} & \textbf{Performance} ($\uparrow$) & \textbf{Sample Efficiency} ($\downarrow$) & \textbf{Performance} ($\uparrow$) & \textbf{Sample Efficiency} ($\downarrow$) \\
    \midrule
    \multirow{3}{*}{\shortstack{DQN+IQN\\\texttt{(DoorKey)}}}
     & {8$\times$8} & \res{\zerov{0\%}}{1.00}{1.00} & \res{\best{22\%}}{139K}{178K} & \res{\zerov{0\%}}{1.00}{1.00} & \res{\best{26\%}}{139K}{187K} \\
     & {12$\times$12} & \res{\best{61\%}}{1.00}{0.62} & \res{\best{57\%}}{277K}{640K} & \res{\best{22\%}}{1.00}{0.82} & \res{\best{37\%}}{292K}{467K} \\
     & {16$\times$16} & \res{\best{283\%}}{0.92}{0.24} & \res{\best{53\%}}{420K}{893K} & \res{\best{92\%}}{0.92}{0.48} & \res{\best{51\%}}{305K}{622K} \\
    \midrule
    \multirow{3}{*}{\shortstack{SAC+TD3\\\texttt{(Scene)}}}
     & {3} & \res{\zerov{0\%}}{1.00}{1.00} & \res{\best{40\%}}{213K}{354K} & \res{\zerov{0\%}}{1.00}{1.00} & \res{\best{20\%}}{213K}{266K} \\
     & {4} & \res{\best{22\%}}{1.00}{0.82} & \res{\best{62\%}}{295K}{776K} & \res{\best{2\%}}{1.00}{0.98} & \res{\best{38\%}}{394K}{634K} \\
     & {5} & \res{\best{147\%}}{0.89}{0.36} & \res{\best{55\%}}{366K}{819K} & \res{\best{68\%}}{0.89}{0.53} & \res{\best{37\%}}{508K}{805K} \\
    \bottomrule
    \end{tabular}%
    }
    % \vspace{-6pt}
\end{table}

\subsection{Main Results: VLMs are Useful for RL Data Prioritization}
\label{sec:main_results}

\looseness=-2
We evaluate \fmethod and baselines across both discrete and continuous environments, as detailed in Section~\ref{sec:experimental-setup}.
Table~\ref{tab:experiments-agg-algo} summarizes the aggregate improvements in both performance and sample efficiency.
\fmethod matches or exceeds every baseline in success and improves sample efficiency in 23 of 24 comparisons.
The largest gains appear in the most challenging settings: on \texttt{DoorKey-16x16}, \fmethod achieves a \textbf{+283.3\%} improvement over \fuer and \textbf{+91.7\%} over \fpert; on \texttt{Scene-5}, the improvement over \fuer is \textbf{+147.2\%} and \textbf{+67.9\%} over \fpert.
Even in cases where baselines eventually solve the task, \fmethod achieves substantially better sample efficiency, reducing the number of steps needed to reach baseline performance by as much as \textbf{62.0\%}.

\begin{wrapfigure}{r}{0.5\linewidth}
    \vspace{-12pt}
    \centering
    \includegraphics[width=\linewidth]{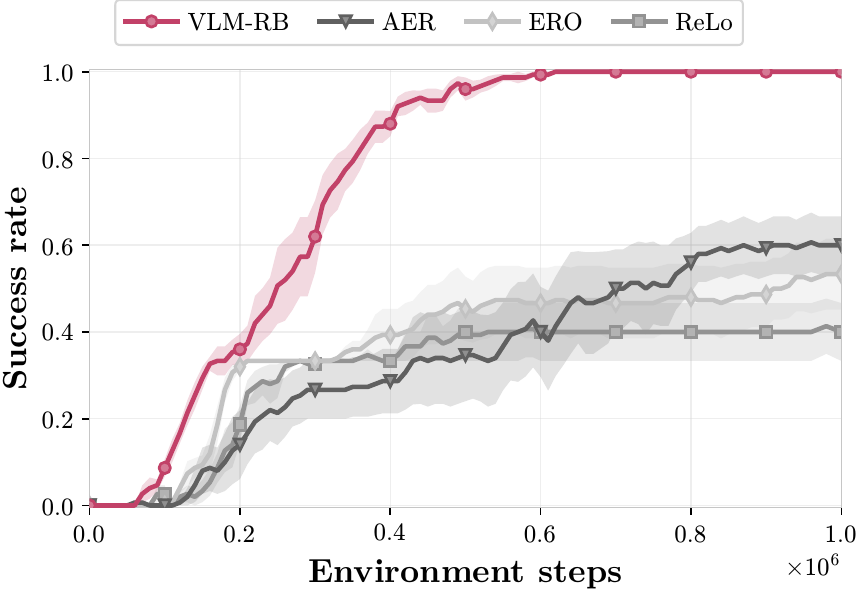}
    \caption{\looseness=-2 \fmethod \textbf{overcomes the failure modes of heuristic prioritization in sparse-reward tasks.} Aggregated success rates on \texttt{MiniGrid/DoorKey} across grid sizes. While alternative methods (AER, ERO, ReLo) fail as they depend on dense rewards or local similarity metrics, \fmethod successfully bridges the long-horizon dependencies. Curves show the mean across 5 seeds, with shaded standard errors.}
    \label{fig:baselines}
    \vspace{-8pt}
\end{wrapfigure}

\textbf{Comparison to Baselines.}
\looseness=-2
Figure~\ref{fig:baselines} expands this comparison, showing that \fmethod is the only method to consistently solve the task when compared to alternative prioritization schemes (AER, ERO, ReLo).
The baselines struggle primarily due to the combination of sparse rewards and long sequential dependencies. ERO and ReLo fail to prioritize task-relevant transitions because their feedback signals are either dominated by noise or remain zero throughout the long pre-success phase, while similarity-based methods like AER over-sample current states and neglect critical past dependencies (such as key acquisition), impeding temporal credit assignment over long horizons.
Further analysis is provided in Appendix~\ref{app:baselines}.

\textbf{Training Time.}
\looseness=-2
Finally, we address the computational trade-off. Although VLM inference introduces a moderate throughput overhead (approximately 12\% wall-clock time per step, see Appendix~\ref{app:throughput-ablation}), this cost is outweighed by the reduction in training steps. In the vast majority of configurations (23/24), \fmethod reduces the absolute wall-clock time to reach peak performance by up to \textbf{67\%}, with speedups exceeding \textbf{20--40\%} in the more challenging exploration settings; the sole exception is a 23\% slowdown (full results in Table~\ref{tab:experiments-full}, Appendix~\ref{app:experiments}).

\subsection{Which Design Choices Matter? Mixing and Scale}
\label{sec:ablations}

\textbf{Sampling Mixture.}
A central component of our approach is to mix VLM-scored samples with uniformly sampled data, allowing the external semantic prior to guide replay while preserving uniform coverage.
To assess its importance, we vary the maximal mixing coefficient, $\lambda_{\max}$, holding all other parameters fixed.
The no-mixture configuration fails to solve the task, whereas the scheduled configurations succeed, showing the importance of mixing VLM-prioritized and uniform replay during training; Appendix~\ref{app:mix-schedule-ablation} provides the full analysis, and Appendix~\ref{app:sampling-frequencies} reports the realized sampling frequencies under the default $\lambda_{\max}=0.5$ mixture.
Appendix~\ref{app:label-corruption} directly tests this safeguard under controlled VLM label corruption.

\textbf{VLM Size.}
\looseness=-2
We also ablated the VLM size and found that scaling the model beyond 1B parameters dramatically increases computational cost without yielding consistent improvements in downstream RL performance (see Appendix~\ref{app:vlm-size-ablation} for the trade-offs between throughput, memory, and performance).

\section{Related Work}\label{sec:related}

\textbf{Prioritized Experience Replay.}
\looseness=-2
Extensive research has focused on defining \emph{which} transitions maximize learning utility.
The seminal work, \fpert~\citep{schaul2015prioritized}, prioritizes transitions with high TD error, serving as a proxy for ``surprising'' events.
Subsequent methods have proposed alternative utility metrics based on state similarity~\citep{sun2020attentive}, potential loss reduction~\citep{sujit2023prioritizing}, target quality/discrepancy criteria~\citep{kumar2020discor}, or by explicitly learning a replay policy~\citep{zha2019experience}.
However, these approaches typically derive priorities from internal, training-dependent signals (e.g., value estimates or TD errors), which can be noisy or undefined early in training.
In contrast, \fmethod leverages a pre-trained VLM to assign semantic priorities, yielding a robust utility score from the very first update.
\looseness=-2
Orthogonal to the choice of \emph{what} to prioritize, several works modify \emph{how} transitions are sampled.
~\citet{hong2022topological} argue for enforcing the ordering of updates within a collected episode, to better align with the Bellman loss that is common in off-policy learning.
Another example, ~\citet{lahire2021large}, focuses on the ``staleness'' of prioritization scores and suggests a double-sampling approach.
These directions are largely complementary to \fmethod and can be combined with VLM-based priorities. Furthermore, because \fmethod computes priorities from a frozen VLM, it naturally avoids the score staleness issue addressed by the latter.
\looseness=-2
A distinct line of work instead alters the \emph{optimization objective} itself, deriving reweighting schemes from $f$-divergence regularization~\citep{li2024roer}, trust-region constraints~\citep{novati2019remember}, or density-ratio estimation~\citep{liu2021regret, sinha2022experience}.
\fmethod changes no loss function, and so remains modular and compatible with a wide range of existing algorithms.

\textbf{LLM/VLM-Enhanced RL.}
Recent work has increasingly used foundation models to improve RL by injecting semantic knowledge into the learning loop.
Broadly, these methods leverage LLMs/VLMs to (i) define richer learning signals (reward supervision),
(ii) propose plans and exploration targets (high-level guidance), or (iii) assist action selection (policy components). Comprehensive surveys can be found in~\citet{cao2024survey} and \citet{schoepp2025evolving}.
\looseness=-2
A major thread treats LLMs and VLMs as sources of reward.
Off-the-shelf VLMs have been used as zero-shot reward scorers to judge state--goal alignment~\citep{rocamonde2023vision};
LLMs can synthesize dense shaping functions from natural-language task descriptions~\citep{xie2023text2reward};
and several methods replace or augment human feedback by using VLM/LLM judgments (absolute ratings or pairwise preferences) to train reward models~\citep{fu2024furl, luu2025enhancing, singh2025varp, ghosh2025preference}.
Taken together, these approaches reframe reward design as semantic labeling, allowing pre-trained models to provide informative supervision in settings where environment rewards are sparse or absent.
A second strand uses foundation models for planning and exploration.
LLMs decompose instructions into temporally-extended plans grounded in affordances~\citep{ahn2022can, huang2022language},
and guide exploration by proposing semantically diverse subgoals or novelty-oriented objectives using learned embeddings~\citep{ma2025explorllm, gupta2022foundation}.
This high-level guidance is particularly useful in settings where unguided exploration rarely visits meaningful states.
A third direction integrates LLMs/VLMs more directly into control, either as action advisors or as policy components.
Examples range from generating executable policy code~\citep{liang2022code} to large-scale vision--language--action (VLA) models that map multimodal inputs to actions~\citep{zitkovich2023rt}.
\looseness=-2
Despite this breadth, most prior work uses these models to shape what the agent optimizes (rewards), where it searches (plans and exploration), or how it acts (policy),
which commonly introduces additional training stages, fine-tuning, or computationally expensive inference.
In contrast, \fmethod targets the replay buffer---the core mechanism enabling sample reuse in off-policy RL---a component which remains underexplored for semantic foundation-model supervision.
Moreover, as our experiments show, open-source pre-trained VLMs can already yield consistent gains when used to prioritize replay, without any additional model training or fine-tuning.

\section{Conclusions}
\label{sec:conclusion}

In this work, we presented \fmethod, a framework that integrates the semantic reasoning capabilities of pre-trained VLMs directly into the experience replay mechanism.
By shifting prioritization from statistical proxies such as TD-error to semantic evaluations of task progress, \fmethod effectively identifies and promotes meaningful experiences even in sparse-reward regimes where traditional metrics fail.
Our empirical results demonstrate that this approach significantly improves both sample efficiency and asymptotic performance across discrete and continuous domains, without requiring fine-tuning or gradient updates to the VLM.

\textbf{Limitations.}
While our approach yields substantial improvements, it is important to clarify the specific assumptions and constraints under which these gains are realized.
First, \fmethod fundamentally relies on the assumption that task progress can be reliably inferred from visual observations.
As a result, \fmethod does not apply to domains where the underlying state lacks a visual representation, such as non-spatial biological systems or abstract network control tasks. In these settings, the VLM cannot ground its reasoning.
Second, querying the VLM introduces additional computational overhead.
Empirically, we found that this overhead reduced throughput by approximately $12\%$ in our experiments.
It is important to note that this computational cost is not intrinsic to the method; the precise overhead depends on hardware configuration and implementation choices, so the trade-off may vary across different setups\footnote{Our implementation did not employ aggressive optimization techniques (e.g., quantization or dedicated serving stacks), suggesting that this overhead could be further minimized.}.
Finally, while our approach uses a goal-agnostic, domain-generic prompt, performance is ultimately bound by the VLM's zero-shot visual reasoning capabilities, which may degrade in visually unstructured environments or require more careful prompt engineering and score calibration.

\textbf{Future Work.}
Our findings suggest several concrete directions for future research.
A natural extension is the application of \fmethod to Goal-Conditioned RL \citep{schaul2015universal}, where the textual description of the current goal can be injected into the VLM prompt to dynamically score alignment, or hindsight textual goals could promote meaningful sub-goals which may have some semantic relation but no direct relevance to the current task~\citep{luu2021hindsight, jiang2019language}.
Another promising direction is to use the replay buffer as a curriculum mechanism, by evolving the prompt over the course of training, prioritizing specific skills such as ``open drawer'' in the early stages and later shifting focus to more compositional behaviors.
Finally, the prioritization scheme could be extended beyond single VLM scoring, for example, by incorporating more robust inference techniques such as majority voting or LLM-as-a-Judge.

\subsection*{AI use statement}

Generative AI tools were used to assist with planning and reviewing ablation experiments, editing analysis and plotting code, organizing and reformatting experimental artifacts, checking numerical consistency, and revising the manuscript for clarity and formatting. Experimental observations were produced by the reported runs, not generated by AI. The authors inspected and tested AI-assisted code, verified reported numbers and claims against raw logs and exported artifacts, and reviewed all AI-assisted text and figures. The authors take responsibility for the final content of this work.

\subsection*{Ethics statement}
This work advances sample-efficient reinforcement learning in domains where visual priors and semantic reasoning offer a tangible advantage by leveraging vision-language models to prioritize experiences. Like other improvements to RL training pipelines, the technique we introduce is method-level and not tied to a specific deployment; it could, in principle, accelerate beneficial applications (e.g., robotics for assistive tasks) or, like any RL advance, downstream applications with negative externalities. We do not foresee specific risks beyond those generally associated with RL research.

{
\small
\bibliographystyle{iclr2027_conference}
\bibliography{iclr2027_conference}
}

\newpage
\appendix

\section{\fmethod}

\begin{algorithm}[H]
\caption{\flare{4}{VLM-prioritized replay buffer}}
\label{alg:vlmrb-async}
\begin{algorithmic}[1]
\footnotesize

\Require prompt $\mathsf{P}$; clip length $L$.
\Statex replay buffer $\gD$; clip buffer $\gC$; queues $\gQ_\mathrm{in}$ and $\gQ_\mathrm{out}$.

\BeginBox[
  fill=flare4!8,
  draw=flare4!40,
  rounded corners=2pt,
  inner sep=2pt
]
\Function{\textsc{VLMWorker}}{}
    \While{true}
        \State $(\texttt{idxs}, \tau^O) \gets \textbf{pop}(\gQ_{\mathrm{in}})$
        \State \makebox[\linewidth][l]{
            $\bm{\mathrm{p}}^{\scriptscriptstyle\mathrm{VLM}}=f_{\scriptscriptstyle\mathrm{VLM}}(\tau^O, \mathsf{P})\in\mathbb{R}$~
            \hfill{\scriptsize\color{gray}(\Eqref{eq:vlm_scoring})}
        }
        \State \textbf{push}$(\gQ_{\mathrm{out}}, (\texttt{idxs}, \bm{\mathrm{p}}^{\scriptscriptstyle\mathrm{VLM}}))$
    \EndWhile
\EndFunction
\EndBox

\vspace{4pt}

\State \textbf{Init:} launch \textsc{VLMWorker} asynchronously.
\While{\textsc{Training}}
    \LComment{\color{flare4}(1) Env step}
    \State Step env to obtain $(s,a,r,s')$ and visual observation $o=\psi(s)$
    \State $\texttt{idx} \gets \textsc{Insert}(\gD,(s,a,r,s' ; \bm{\mathrm{p}}^{\scriptscriptstyle\mathrm{VLM}}\gets\bar{\bm{\mathrm{p}}}))$
    \State $\gC \gets \gC \cup (\texttt{idx},o)$

    \LComment{\color{flare4}(2) Enqueue clips (streaming)}
    \If{$|\gC| = L$ \textbf{or} $\mathtt{terminated}$ \textbf{or} $\mathtt{truncated}$}
        \State $\texttt{idxs}, \tau^O \gets \{(\texttt{idx}_i, o_i)\}_{i=0}^{|\gC|-1}$
        \State \textbf{push}$(\gQ_{\mathrm{in}}, (\texttt{idxs}, \tau^O))$
        \State $\gC \gets \emptyset$
    \EndIf

    \LComment{\color{flare4}(3) Apply VLM scores (drain output queue)}
    \While{$\gQ_{\mathrm{out}}$ not empty}
        \State $(\texttt{idxs}, \bm{\mathrm{p}}^{\scriptscriptstyle\mathrm{VLM}}) \gets \textbf{pop}(\gQ_{\mathrm{out}})$
        \ForAll{$\texttt{idx} \in \texttt{idxs}$}
            \State \textsc{SetVLMScore}$(\gD,\texttt{idx},\bm{\mathrm{p}}^{\scriptscriptstyle\mathrm{VLM}})$
        \EndFor
        \State $\bar{\bm{\mathrm{p}}} \gets \textsc{CMA}(\bar{\bm{\mathrm{p}}}, \bm{\mathrm{p}}^{\scriptscriptstyle\mathrm{VLM}})$
    \EndWhile

    \LComment{\color{flare4}(4) Sample and learn}
    \State \makebox[\linewidth][l]{
    $\gB \sim q_t$
    \hfill{\scriptsize\color{gray}(\Eqref{eq:sampling_mixture})}
    }
    \State \textsc{UpdateLearner}$(\gB)$; update $\lambda_t$
\EndWhile

\end{algorithmic}
\end{algorithm}

In Algorithm~\ref{alg:vlmrb-async}, $\mathcal{C}$ denotes a temporary clip buffer, $\mathcal{Q}_{\text{in}}$ and $\mathcal{Q}_{\text{out}}$ are asynchronous communication queues, and CMA refers to the cumulative moving average of returned VLM scores.

\section{Environment Details}
\label{app:environments}

In this section, we describe the environments used in our experiments.

\textbf{MiniGrid / DoorKey.}
We use the \texttt{DoorKey} tasks from the MiniGrid suite~\citep{chevalier2023minigrid} with sizes \texttt{8x8}, \texttt{12x12}, and \texttt{16x16}.
Each episode requires the agent to complete a sequence of subtasks: (i) navigate to and pick up the key, (ii) reach the locked door and open it using the  \texttt{toggle} action (which is only possible while holding the key), and (iii) proceed to the goal tile. This structure enforces temporal dependencies and compositional reasoning.
The reward function is sparse: the agent receives zero reward at all intermediate steps and a positive reward only upon reaching the goal, with the reward magnitude linearly decaying with the number of steps taken, following the standard MiniGrid protocol.
Episodes terminate either upon successful completion or when the maximum step limit is reached.

State observations are provided in a symbolic format, where each $s_t \in \{0,1,\ldots\}^{N \times N \times 3}$ encodes the full grid. The three channels correspond to (i) object indices, (ii) color indices, and (iii) a state channel that includes both the door state and the agent's orientation.
The action space is discrete with $|\mathcal{A}|=5$, consisting of turn left, turn right, move forward, pick up, and toggle actions.

\textbf{OGBench / Scene.}
We use the \texttt{scene-play} manipulation environment from OGBench~\citep{park2024ogbench}, which is a MuJoCo-based tabletop task involving a UR5e arm with a parallel gripper. The agent interacts with a set of objects: a cube, a sliding drawer, a sliding window, and two toggle buttons.
Each button controls the lock state of either the drawer or the window, introducing dependencies that require the agent to execute explicit unlock, manipulate, and (re-)lock sequences to achieve many goals.
Episodes are capped at a horizon of $750$ steps. To facilitate learning, we employ potential-based reward shaping with a per-step living cost, defined as:
\[
r_t = -1 \;+\; \gamma \, \phi(s_{t+1}) - \phi(s_t),
\qquad
\phi(s) = \frac{1}{K}\sum_{i=1}^{K}\mathbb{I} \left[\text{subgoal}_i(s)\right].
\]
Here, $\phi(s)$ denotes the fraction of satisfied subgoals, where each subgoal is a task-specific predicate over cube placement, button states, and the positions of the drawer and window. An episode terminates either when all subgoals are satisfied ($\phi(s_t)=1$) or when the maximum horizon is reached.

Observations are provided as full state vectors, $s_t \in \mathbb{R}^{40}$, and actions are specified as continuous end-effector controls, $a_t \in \mathbb{R}^{5}$.

To address the exploration challenges inherent in these long-horizon tasks, we initialize the replay buffer with a lightweight warm-start of $10$ demonstration episodes, which are held fixed across all methods. This small set of demonstrations provides minimal task-relevant coverage, while leaving the remainder of the training protocol unchanged.
Warm-starting the buffer this way follows common practice in demonstration-guided RL~\citep{hester2018deep, vecerik2017leveraging, nair2018overcoming}. All methods share the same $10$ episodes, task reward, and shaping; only the replay priority differs. \texttt{MiniGrid/DoorKey} uses no demonstrations.

We evaluate on a set of predefined Scene goals (tasks 3--5), which are designed to progressively increase the degree of temporal composition required for successful completion.
\begin{description}
    \item[\textbf{Task 3 (rearrange-medium).}]
    The agent must move the cube to a specified tabletop location, open the drawer, close the window, and terminate with both the drawer and window unlocked. Notably, the window begins in a locked and open state, so the policy must first unlock it before closing, while simultaneously coordinating the manipulation of the drawer and the relocation of the cube.

    \item[\textbf{Task 4 (put-in-drawer).}]
    The agent is required to place the cube inside the drawer and terminate with the drawer closed and unlocked, while ensuring the window remains locked. This sequence involves unlocking the drawer, opening it, inserting the cube, and then closing the drawer.

    \item[\textbf{Task 5 (rearrange-hard).}]
    The agent must place the cube inside the (closed) drawer and leave the window open, while ensuring that both the drawer and window are locked at the end of the episode. Achieving this goal requires the agent to execute both unlock and relock sequences, coordinate drawer opening and closing, and correctly position the cube.
\end{description}

\newpage

\section{Prompts}
\label{app:prompts}

\textbf{Task prompts used in our experiments.}
We use a binary ``success visible'' query with a strict output format.
Both prompts are \emph{goal-agnostic}: neither names the goal, the objects involved, or any success predicate, so the same text is reused unchanged across all \texttt{DoorKey} grid sizes ($8\times8$, $12\times12$, $16\times16$) and across \texttt{Scene} tasks $3$--$5$, whose goals differ.
They are also \emph{domain-generic} in the sense that the only domain-dependent content is the list of visual cues in the \texttt{Scene} variant, which describes manipulation in general rather than any particular task.
We report the first phrasing we wrote in each case; neither was tuned against downstream RL performance, and no prompt selection was performed.

\begin{enumerate}[leftmargin=*, topsep=2pt, itemsep=2pt]
    \item \texttt{MiniGrid/DoorKey}:
    \emph{``Does this clip contain a clear instance of goal satisfaction anywhere in it? If no visible success occurs, answer No. Do not guess. Output exactly Answer: Yes or Answer: No.''}

    \item \texttt{OGBench/Scene}:
    \emph{``Is there at least one clear instance of goal satisfaction in these frames? Look for contact + displacement consistent with the goal (lift off surface, place into receptacle, open/close articulation, move to target zone). Do not guess. If not visible, answer No. Output exactly Answer: Yes or Answer: No.''}
\end{enumerate}

\textbf{Meta-prompt for generating task prompts.}
To facilitate applying our procedure to new environments, we use the following meta-prompt to generate a task prompt from a task identifier and optional human context. The meta-prompt encourages generic, visually grounded success criteria and enforces a strict output schema.

\begin{promptbox}[Meta-prompt template for generating task prompts]
\user{
You are an expert in Reinforcement Learning and Visual Language Models. I will provide a short clip from an agent rollout. Your job is to write a single text prompt that asks a VLM to output a \textbf{binary} judgment about whether \textbf{goal satisfaction / competent task progress is clearly visible} in the clip.

\textbf{Requirements:}
\begin{itemize}[leftmargin=*, topsep=2pt, itemsep=2pt]
    \item Keep the prompt \textbf{environment-agnostic}: rely on visible physics and outcomes rather than simulator-specific rules.
    \item Specify what to look for in broad categories when helpful (e.g., contact + displacement for manipulation; reaching a target region for navigation).
    \item Explicitly instruct: \textbf{do not guess}; if success is not clearly visible, answer No.
    \item End the prompt with the string: \textbf{Output exactly Answer: Yes or Answer: No.}
\end{itemize}

\textbf{Examples (few-shot):}
\begin{itemize}[leftmargin=*, topsep=2pt, itemsep=2pt]
\item \textbf{Task:} Minigrid-doorkey \\
\textbf{Prompt:} ``Does this clip contain a clear instance of goal satisfaction anywhere in it? If no visible success occurs, answer No. Do not guess. Output exactly Answer: Yes or Answer: No.''
\item \textbf{Task:} OGBench-scene-task-4 \\
\textbf{Prompt:} ``Is there at least one clear instance of goal satisfaction in these frames? Look for contact + displacement consistent with the goal (lift off surface, place into receptacle, open/close articulation, move to target zone). Do not guess. If not visible, answer No. Output exactly Answer: Yes or Answer: No.''
\end{itemize}

\textbf{Current request:}\\
\textbf{Task Name:} \texttt{\{\{TASK\_ID\}\}}\\
\textbf{Optional Human Context:} \texttt{\{\{HUMAN\_CONTEXT\}\}}\\
}
\end{promptbox}

\textbf{Example: prompt families for a new benchmark.}
As an illustration, we can apply the meta-prompt to a suite such as DeepMind Control by grouping tasks by their visible success patterns (locomotion / balance / interaction) and instantiating one generic prompt per group.

\begin{promptbox}[Example: prompt families for DeepMind Control Suite]
\user{
Group DeepMind Control tasks by shared \emph{visible} success patterns and propose one generic success-detection prompt per group (not task-specific). Each prompt must end with: ``Output exactly Answer: Yes or Answer: No.''
}
\model{
    We used Gemini 3 Pro to generate these prompts.
    Here is the partition into three semantic groups based on the \textbf{nature of the movement}, with a generalized ``vibe-check'' prompt for each.

    \vspace{1em}
    \noindent\textbf{Group 1: Going Somewhere (Locomotion)}

    \textbf{The Semantic Goal:} The agent must make progress across the environment. The VLM needs to distinguish between an agent that is struggling/falling (chaos) and one that is moving purposefully (control).

    \vspace{0.5em}
    \noindent\textbf{The Group:}
    \texttt{cheetah-run}, \texttt{dog-walk}, \texttt{dog-trot}, \texttt{dog-run}, \texttt{fish-swim}, \texttt{hopper-hop}, \texttt{humanoid-walk}, \texttt{humanoid-run}, \texttt{quadruped-walk}, \texttt{quadruped-run}, \texttt{walker-walk}, \texttt{walker-run}.

    \vspace{0.5em}
    \noindent\textbf{The Prompt:}
    \begin{quote}
    ``Does the agent demonstrate controlled, purposeful movement through the environment? Look for the agent making progress in a specific direction (locomotion) rather than just flailing in place, falling over, or crashing. The movement should look stable and competent. If the agent falls, flips, or fails to move effectively, answer No. Output exactly Answer: Yes or Answer: No.''
    \end{quote}

    \noindent\rule{\linewidth}{0.4pt}

    \vspace{1em}
    \noindent\textbf{Group 2: Defying Gravity (Stabilization)}

    \textbf{The Semantic Goal:} The agent is fighting gravity. Success is defined by \textbf{stillness} or \textbf{verticality}. Whether it is a pole swinging up or a dog standing still, the ``vibe'' of success is reaching a stable, upright state and holding it.

    \vspace{0.5em}
    \noindent\textbf{The Group:}
    \texttt{acrobot-swingup}, \texttt{cartpole-balance}, \texttt{cartpole-balance\_sparse}, \texttt{cartpole-swingup}, \texttt{cartpole-swingup\_sparse}, \texttt{dog-stand}, \texttt{hopper-stand}, \texttt{humanoid-stand}, \texttt{walker-stand}, \texttt{pendulum-swingup}.

    \vspace{0.5em}
    \noindent\textbf{The Prompt:}
    \begin{quote}
    ``Does the agent successfully achieve and maintain a stable, upright posture against gravity? Look for the agent balancing itself or an object vertically. Success means the agent (or the object it controls) is standing up or swinging up to a balanced position and staying there without collapsing, spinning out of control, or falling over. If it looks unstable or falls, answer No. Output exactly Answer: Yes or Answer: No.''
    \end{quote}

    \noindent\rule{\linewidth}{0.4pt}

    \vspace{1em}
    \noindent\textbf{Group 3: Touching the Thing (Interaction)}

    \textbf{The Semantic Goal:} This is about precise manipulation. The agent is not trying to go somewhere or just stand up; it is trying to force an interaction between its end-effector and a target object.

    \vspace{0.5em}
    \noindent\textbf{The Group:}
    \texttt{ball\_in\_cup-catch}, \texttt{finger-spin}, \texttt{finger-turn\_easy}, \texttt{finger-turn\_hard}, \texttt{reacher-easy}, \texttt{reacher-hard}.

    \vspace{0.5em}
    \noindent\textbf{The Prompt:}
    \begin{quote}
    ``Does the agent successfully interact with the target object in this scene? Look for the agent reaching a target, catching an object, or spinning a mechanism as intended. There should be clear, controlled contact or manipulation consistent with solving a task. If the agent misses the target, drops the object, or flails without making meaningful contact, answer No. Output exactly Answer: Yes or Answer: No.''
    \end{quote}
    }
\end{promptbox}

\newpage

\section{Baselines}
\label{app:baselines}

\texttt{DoorKey} presents a particularly challenging regime, characterized by extremely sparse rewards (identically zero until termination) and a long-horizon, sequential dependency structure (key to door to goal).
In such settings, replay schemes relying on dense feedback, informative TD statistics, or a well-defined similarity metric are fundamentally misaligned with the task: they fail to prioritize transitions relevant to solving the task, particularly throughout the extended pre-success phase.

\textbf{Experience Replay Optimization (ERO)~\citep{zha2019experience}.}
ERO optimizes a learned rejection sampler using a REINFORCE-style update, where the scalar reward is defined as the change in evaluation return.
In \texttt{DoorKey}, evaluation returns remain identically zero for a substantial portion of training (reflecting the absence of successes), which in turn implies $r_{\mathrm{replay}} \approx 0$ and results in near-zero gradients for the replay policy over extended periods.
Furthermore, the replay-policy features used here, $(r_i,|\delta_i|, t_i/T_{\max})$, are weakly informative under sparse rewards: $r_i=0$ for almost all transitions and $|\delta_i|$ is largely determined by bootstrap noise in the early stages of learning.
As a result, the rejection policy effectively acts as an untrained stochastic filter for much of the training.
When successes eventually occur, $r_{\mathrm{replay}}$ becomes highly variable, leading to unstable and non-stationary updates which can bias acceptance toward incidental correlates, such as late timesteps.
The top-up rule, which fills with the highest-$P_i$ candidates when acceptance is low, further concentrates replay on a narrow subset of transitions. This reduces diversity and fails to provide a consistent signal for solving the task.

\textbf{ReLo (Reducible Loss Prioritization)~\citep{sujit2023prioritizing}.}
ReLo prioritizes transitions based on the difference between the magnitudes of online and target TD residuals, $\bm{\mathrm{p}}_{\mathrm{ReLo}} = \max(0, |\delta_{\theta}| - |\delta_{\theta^-}|) + \varepsilon$.
In the sparse-reward \texttt{DoorKey} setting, both residuals are typically dominated by bootstrapping error rather than meaningful reward propagation throughout the prolonged pre-success phase.
Moreover, the use of hard target updates periodically synchronizes the online and target networks, further diminishing any systematic separation between $|\delta_{\theta}|$ and $|\delta_{\theta^-}|$.
Consequently, $|\delta_{\theta}| - |\delta_{\theta^-}|$ is often small or negative and is clipped to approximately $\varepsilon$, so the sampling distribution effectively reverts toward uniform, yet still incurs the variance and bias tradeoffs associated with prioritized replay.

\textbf{Attentive Experience Replay (AER)~\citep{sun2020attentive}.}
AER selects samples closest (in a frozen embedding space) to the agent's current state, thereby inducing a strongly local replay distribution.
In \texttt{DoorKey} this notion of locality is fundamentally misaligned with the task structure. Once trajectories reach later phases (near the door or goal), nearest-neighbor replay disproportionately samples those regions and neglects earlier structural dependencies, such as key acquisition, thereby impeding temporal credit assignment across the long horizon.
This effect is further amplified by the use of a randomly initialized frozen encoder: the resulting metric primarily captures superficial spatial proximity in the grid encoding, rather than functional or task-relevant similarity.
As a result, the selection rule acts as a myopic location-based filter, rather than an attentive semantic sampler.

\newpage
\section{Implementation Details}
\label{app:impl-details}

To ensure a fair comparison, all baselines are implemented with the same backbone architecture, optimizer, and training schedule as \fmethod.
Unless otherwise specified, each method employs a \fpert-style replay buffer parameterized by exponents $\alpha$ and $\beta$.

\subsection{Baseline Implementations}

\textbf{Uniform Experience Replay.}
\label{app:uniform}
In this setting, each transition $i$ in the buffer is sampled with equal probability, $q_t^{\mathrm{U}}(i) = 1/N_{\text{curr}}$, where $N_{\text{curr}}$ denotes the current buffer size.

\textbf{\fpert (Prioritized Experience Replay).}
\label{app:per}
Here, the priority assigned to each transition $i$ is given by $\bm{\mathrm{p}}_i = |\delta_i| + \varepsilon$, where $\delta_i$ denotes the most recent TD-error for that transition, and $\varepsilon=10^{-6}$ guarantees that every transition can be sampled.
The probability of sampling transition $i$ is then defined as $$q_t^{\mathrm{P}}(i) = \frac{\bm{\mathrm{p}}_i^\alpha}{\sum_k \bm{\mathrm{p}}_k^\alpha},$$ with $\alpha$ determining the extent to which prioritization influences sampling.
To account for the bias from prioritized sampling, importance-sampling (IS) weights are computed as $$w_i = \left( \frac{1}{N} \cdot \frac{1}{q_t^{\mathrm{P}}(i)} \right)^\beta.$$ These weights are normalized by $1/\max_j(w_j)$ prior to being used in the loss.

\textbf{Experience Replay Optimization (ERO).}
\label{app:ero}
ERO~\citep{zha2019experience} replaces standard replay sampling with a learned rejection policy applied to uniformly drawn candidates.
The replay policy is parameterized as an MLP $\phi_\psi$ with two hidden layers of 64 units each (ReLU activations, sigmoid output), mapping transition features to a retention probability.
$$\bm{\mathrm{p}}_i=\phi_\psi\!\left(r_i,\;|\delta_i|,\;t_i/T_{\max}\right)\in(0,1),$$
where $r_i$ is the immediate reward, $|\delta_i|$ is the absolute TD-error stored in the buffer (refreshed after each gradient step), and $t_i/T_{\max}$ is the normalized within-episode timestep.
For each new transition, the absolute TD-error is initialized to $|\delta|=1.0$ upon insertion.
At each update, a candidate pool of size $N_{\text{cand}}=4B$ is drawn uniformly from the buffer, where $B$ is the batch size. Each candidate is independently retained with probability $\bm{\mathrm{p}}_i$. If fewer than $B$ transitions are accepted, the batch is completed by selecting the remaining candidates with the highest $\bm{\mathrm{p}}_i$, ensuring a batch of size $B$. As in \citet{zha2019experience}, importance sampling is not applied.
The replay policy is updated using a REINFORCE-style gradient, where evaluation returns serve as the reward signal.
Let $\bar{R}$ represent a moving average of the evaluation return. After each evaluation, the reward for the replay policy is computed as $r_{\text{replay}}=\bar{R}_{\text{current}}-\bar{R}_{\text{previous}}$, and $\bar{R}$ is updated accordingly.
With $\mathcal{B}$ denoting the selected batch, the replay policy loss is$$\mathcal{L}_{\text{ERO}}=-\, r_{\text{replay}} \sum_{i\in \mathcal{B}} \log \bm{\mathrm{p}}_i,$$where gradients are applied only to $\psi$.

\textbf{ReLo (Reducible Loss Prioritization).}
\label{app:relo}
ReLo~\citep{sujit2023prioritizing} assigns priority to each transition based on a reducible-loss score, which is computed from the online and target Q-networks.
Given a transition $(s,a,r,s')$, the target value is defined as $y=r+\gamma\max_{a'}Q_{\theta^-}(s',a')$. The online and target TD-errors are then $\delta_{\text{online}}(s,a)=Q_{\theta}(s,a)-y$ and $\delta_{\text{target}}(s,a)=Q_{\theta^-}(s,a)-y$, respectively.
The sampling priority is
$$\bm{\mathrm{p}}_{\text{ReLo}}(s,a)=\max\!\left(0,\,|\delta_{\text{online}}(s,a)|-|\delta_{\text{target}}(s,a)|\right)+\varepsilon,
\qquad \varepsilon=10^{-6}.$$
Each new transition is initialized with the maximum priority value among all transitions currently in the buffer.
The method sets $\alpha=0.6$ and initializes $\beta=0.4$, annealing $\beta$ linearly to $1.0$. Standard importance-sampling weights are applied.

\textbf{Attentive Experience Replay (AER).}
\label{app:aer}
AER~\citep{sun2020attentive} selects transitions using an attentive sampling mechanism.
A dedicated encoder $\phi$ is constructed with the same backbone architecture as the Q-network. This encoder is randomly initialized and remains fixed throughout training.
At each training step $t$, a candidate pool of size $N_{\text{cand}}=\lfloor \lambda_t B\rfloor$ is sampled uniformly from the buffer, where $\lambda_t$ decays linearly from $\lambda_0=4$ to $1$ over the course of training.
If the candidate pool size $N_{\text{cand}}$ is less than or equal to the batch size $B$, the method reverts to the default sampling strategy.
Given the current state $s_{\text{curr}}$ and a set of candidate states $\{s_i\}$, distances are computed in the frozen embedding space as $d(s_{\text{curr}},s_i)=\|\phi(s_{\text{curr}})-\phi(s_i)\|_2^2$. The $B$ candidates with the smallest distances are then selected deterministically to form the training batch.
Following~\citet{sun2020attentive}, importance sampling is disabled.

\subsection{Agent Architectures}

\textbf{MiniGrid agents (DQN / IQN).}
Both DQN and IQN are trained on symbolic Minigrid observations $s_t \in \{0,\dots\}^{N\times N\times 3}$, which encode object identities, colors, and state information such as door status and agent orientation. To process these inputs, we employ a shared encoder which first embeds each channel, then applies a residual dilated CNN with three blocks. The resulting features are aggregated using both average and max global pooling, yielding a 256-dimensional representation. This design aims to capture both local and global spatial structure in the environment.
For DQN, the 256-dimensional feature vector is passed through a small MLP, which outputs scalar Q-values for each discrete action.
In IQN, we replace the scalar output head with an implicit quantile head. Specifically, we embed sampled quantile fractions using cosine functions, project these embeddings to match the feature dimension, and combine them element-wise with the state features. The resulting representations are mapped to per-action quantile values. We optimize using the quantile Huber loss and select actions using either Double-DQN-style~\citep{van2016deep} or target-network-based strategies, as described in Appendix~\ref{app:impl-details}. This approach allows IQN to model the full distribution over returns, rather than just the mean.

\textbf{OGBench Scene agents (SAC / TD3).}
For \texttt{OGBench/Scene}, both SAC and TD3 operate on flattened state-based observations $s_t \in \mathbb{R}^{40}$. We adopt a unified architecture across tasks: the actor is a fully-connected network with three hidden layers of width 512 and ReLU activations. This network outputs either a tanh-squashed Gaussian policy (for SAC) or a deterministic tanh policy (for TD3), enabling flexible action selection in continuous spaces.
The critic is implemented as an ensemble of $N=10$ Q-functions, each parameterized by a 3-layer MLP that receives the concatenated state and action as input and outputs a scalar value. For target computation, we randomly select $M=2$ ensemble members and aggregate their predictions using a mean for SAC and a minimum for TD3. This ensemble approach is intended to improve stability and reduce overestimation bias.
For actor updates, we use the same Q-function ensemble, aggregating Q-values according to the actor reduction rule. In all reported experiments, SAC uses the mean and TD3 the minimum over ensemble members. This ensures consistency between actor and critic updates.

\subsection{Replay and Scoring Implementation}

\textbf{Replay buffer and prioritization.}
All methods are implemented with a two-branch replay ensemble: one prioritized branch and one uniform branch which share an underlying storage, implemented using \texttt{ReplayBufferEnsemble} \citep{bou2023torchrl} with mixture weights controlled by the sampling ratio schedule in Tables~\ref{tab:ogbench_hyperparams}--\ref{tab:minigrid_hyperparams}.
For the \fpert baseline, we assign priorities using the standard PER formula $\bm{\mathrm{p}}_i = |\delta_i| + \varepsilon$, where $\delta_i$ is the TD error. When importance-sampling corrections are enabled, we apply them as usual. New transitions are initially assigned a small default priority, and we update their TD-errors after each gradient step to ensure accurate prioritization.
In \fmethod, we augment each transition with both a TD-based metric and a VLM-based score. Their product defines the priority signal $\bm{\mathrm{p}}_i$, and the standard \fpert{} transformation produces $q_t^{\mathrm{P}}$ (\Eqref{eq:vlm_td_priority}); \Appref{app:priority-variants} ablates the two factors.
A transition enters the buffer with the provisional VLM score $\bar{\bm{\mathrm{p}}}$, the cumulative moving average (CMA) of the VLM scores returned so far, so that a transition awaiting its score uses the current average.
Once the worker returns the score of the clip containing $i$, $\bm{\mathrm{p}}^{\scriptscriptstyle\mathrm{VLM}}_{i}$ overwrites $\bar{\bm{\mathrm{p}}}$. As in \fpert, $|\delta_i|$ is refreshed at every replay of $i$, and the product is formed at sampling time.

\textbf{VLM worker.}
Decoupling scoring from the learner also makes \fmethod agnostic to the policy input modality --- the VLM requires rendered frames, while the policy may operate on arbitrary observation spaces such as low-dimensional states --- and lets a single VLM instance serve multiple parallel environments.
The VLM worker uses Perception-LM-1B~\citep{cho2025PerceptionLM} with a fixed prompt to detect binary success. It processes non-overlapping clips of $L=32$ frames. The clip buffer is flushed at termination or truncation, so clips never cross episode boundaries; a shorter final clip is left-padded. We fix $L=32$ across all experiments and do not tune it.
We define the raw score $r^{\scriptscriptstyle\mathrm{VLM}}$ as the first generated token's ``Yes''/``No'' mass ratio, summing the probability assigned to all ``Yes'' token variants ($m_{\mathrm{Yes}}$) and all ``No'' variants ($m_{\mathrm{No}}$). The binarized instantiation used in the main experiments is
\begin{equation}
    \label{eq:token_mass_ratio}
    \begin{aligned}
    r^{\scriptscriptstyle\mathrm{VLM}}
    &= \frac{m_{\mathrm{Yes}}}{m_{\mathrm{Yes}} + m_{\mathrm{No}}} \in [0,1], \\
    \bm{\mathrm{p}}^{\scriptscriptstyle\mathrm{VLM}}
    = f_{\scriptscriptstyle\mathrm{VLM}}(\tau^O,\mathsf{P})
    &= \mathbf{1}\!\left[r^{\scriptscriptstyle\mathrm{VLM}}>\tau\right],
    \qquad \tau=0.5.
    \end{aligned}
\end{equation}
The ratio itself requires no calibration data, but a hard threshold depends on the score scale of the chosen VLM and prompt. We use $\tau=0.5$ for the default Perception-LM prompt; transferring the hard-threshold variant to another VLM or prompt may require a small labeled calibration corpus, while using $r^{\scriptscriptstyle\mathrm{VLM}}$ directly avoids threshold calibration. \Appref{app:threshold-ablation} evaluates both choices.
The resulting scores are asynchronously streamed back to update the replay buffer, ensuring the main learner is not blocked during this process.

\textbf{Bounded influence of the VLM score.}
Because a fraction $(1-\lambda_t)$ of every batch is drawn uniformly, the VLM changes how often a transition is replayed, never whether it can be.
For any $f_{\scriptscriptstyle\mathrm{VLM}}$, transition $i$ is sampled at
\begin{equation}
    \label{eq:amplification}
    A_t(i) = \frac{q_t(i)}{1/N_t} = (1-\lambda_t) + \lambda_t N_t\, q_t^{\mathrm{P}}(i)
\end{equation}
times its uniform rate, where $N_t$ is the number of transitions in the buffer. The first term is a floor: with $\lambda_{\max}=0.5$, every transition retains at least half its uniform probability regardless of its score.
A binary score makes the role of the numerical floor explicit. Transitions with zero product signal receive prioritized mass $\varepsilon^\alpha/Z_t$, where $Z_t$ is the denominator in \Eqref{eq:vlm_td_priority}, and are therefore amplified slightly above $(1-\lambda_t)$. When this floor is negligible relative to the positive product signals, the $|P_t|$ promoted transitions absorb nearly all of the prioritized branch and their average amplification approaches $(1-\lambda_t) + \lambda_t N_t/|P_t|$, while all other transitions approach $(1-\lambda_t)$.
This is roughly half the factor purely prioritized sampling would apply, and the warm-up keeps it near $1$ early in training, when the buffer is small and errors would be most costly.

\newpage

\subsection{Hyperparameters}

\begin{table}[H]
\centering
\caption{Hyperparameters and implementation details for OGBench experiments. Shared parameters are listed once; divergent parameters are compared side-by-side.}
\label{tab:ogbench_hyperparams}
\vspace{0.2cm}
\begin{tabular}{lcc}
\toprule
\textbf{Hyperparameter} & \textbf{TD3} & \textbf{SAC} \\
\midrule
\multicolumn{3}{l}{\textit{Network Architecture}} \\
Hidden Dimensions & \multicolumn{2}{c}{[512, 512, 512]} \\
Q-Network Layer Norm & True & True \\
\midrule
\multicolumn{3}{l}{\textit{Optimization}} \\
Critic Learning Rate & \multicolumn{2}{c}{$3 \times 10^{-4}$} \\
Actor Learning Rate & $3 \times 10^{-4}$ & $1 \times 10^{-3}$ \\
Alpha Learning Rate & -- & $3 \times 10^{-4}$ \\
Batch Size & \multicolumn{2}{c}{256} \\
Discount Factor ($\gamma$) & \multicolumn{2}{c}{0.99} \\
Target Smoothing ($\tau$) & \multicolumn{2}{c}{0.005} \\
Max Grad Norm & \multicolumn{2}{c}{10.0} \\
Learning Starts & \multicolumn{2}{c}{10,000 steps} \\
\midrule
\multicolumn{3}{l}{\textit{Ensemble \& Update Ratios}} \\
Ensemble Size ($N$) & \multicolumn{2}{c}{10} \\
Subsampled Q-Networks ($M$) & \multicolumn{2}{c}{2} \\
Critic UTD Ratio & \multicolumn{2}{c}{4} \\
Actor UTD Ratio & \multicolumn{2}{c}{2} \\
Target Q Reduction & $\min$ & $\text{mean}$ \\
Actor Q Reduction & $\min$ & $\text{mean}$ \\
\midrule
\multicolumn{3}{l}{\textit{Data \& Replay Buffer}} \\
Replay Buffer Size & \multicolumn{2}{c}{$1 \times 10^6$} \\
Expert Demos & \multicolumn{2}{c}{10} \\
\midrule
\multicolumn{3}{l}{\textit{Algorithm Specifics}} \\
Policy Update Frequency & 2 & 1 \\
Exploration Noise (std) & 0.1 & -- \\
Target Policy Noise (std) & 0.2 & -- \\
Noise Clip & 0.5 & -- \\
Entropy Coeff. ($\alpha$) & -- & 1.0 (Initial) \\
Target Entropy Scale & -- & 0.5 \\
Auto Entropy Tuning & -- & True \\
\midrule
\multicolumn{3}{l}{\textit{\fpert Baseline Settings}} \\
\fpert Alpha ($\alpha$) & \multicolumn{2}{c}{0.7} \\
\fpert Beta ($\beta$) & \multicolumn{2}{c}{1.0} \\
Importance Sampling & \multicolumn{2}{c}{False} \\
\midrule
\multicolumn{3}{l}{\textit{\fmethod Settings}} \\
VLM Model & \multicolumn{2}{c}{Facebook Perception-LM-1B} \\
Priority Signal & \multicolumn{2}{c}{$\bm{\mathrm{p}}^{\scriptscriptstyle\mathrm{VLM}} \cdot |\delta|$} \\
Priority Mode & \multicolumn{2}{c}{Hard Threshold ($>$ 0.5)} \\
Trajectory Length ($L$) & \multicolumn{2}{c}{32 Frames} \\
Importance Sampling & \multicolumn{2}{c}{False} \\
Prioritized Sampling Ratio ($\lambda_0$ and $\lambda_\mathrm{max}$) & \multicolumn{2}{c}{Annealed $0.0 \to 0.5$} \\
Annealing Schedule ($T_{\mathrm{schedule}}$) & \multicolumn{2}{c}{Linear over 500k steps} \\
\bottomrule
\end{tabular}
\end{table}

\begin{table}[H]
\centering
\caption{Hyperparameters and implementation details for MiniGrid experiments. Shared parameters are listed once; divergent parameters are compared side-by-side.}
\label{tab:minigrid_hyperparams}
\vspace{0.2cm}
\begin{tabular}{lcc}
\toprule
\textbf{Hyperparameter} & \textbf{DQN} & \textbf{IQN} \\
\midrule
\multicolumn{3}{l}{\textit{Optimization}} \\
Learning Rate & \multicolumn{2}{c}{$4 \times 10^{-5}$} \\
Batch Size & \multicolumn{2}{c}{128} \\
Discount Factor ($\gamma$) & \multicolumn{2}{c}{0.95} \\
Target Update Frequency & \multicolumn{2}{c}{1,000 steps} \\
Target Update Rate ($\tau$) & \multicolumn{2}{c}{1.0 (Hard Update)} \\
Learning Starts & \multicolumn{2}{c}{500 steps} \\
Train Frequency & \multicolumn{2}{c}{4 steps} \\
Max Grad Norm & \multicolumn{2}{c}{1.0} \\
\midrule
\multicolumn{3}{l}{\textit{Exploration (Epsilon-Greedy)}} \\
$\mathrm{eps}_\mathrm{start}$ & \multicolumn{2}{c}{1.0} \\
$\mathrm{eps}_\mathrm{end}$ & \multicolumn{2}{c}{0.05} \\
Exploration Fraction & \multicolumn{2}{c}{0.5 (First 50\% of training)} \\
\midrule
\multicolumn{3}{l}{\textit{Algorithm Specifics}} \\
Double DQN & \multicolumn{2}{c}{Enabled} \\
Noisy Nets & -- & Disabled \\
Num Quantiles (Policy) & -- & 32 \\
Num Quantiles (Train/Target) & -- & 64 \\
Num Cosine Basis Functions & -- & 64 \\
Huber Kappa ($\kappa$) & -- & 1.0 \\
\midrule
\multicolumn{3}{l}{\textit{\fpert Baseline Settings}} \\
\fpert Alpha ($\alpha$) & \multicolumn{2}{c}{0.7} \\
\fpert Beta ($\beta$) & \multicolumn{2}{c}{1.0} \\
Importance Sampling & \multicolumn{2}{c}{True} \\
Prioritized Sampling Ratio & \multicolumn{2}{c}{1.0 (Always Prioritized)} \\
\midrule
\multicolumn{3}{l}{\textit{\fmethod Settings}} \\
VLM Model & \multicolumn{2}{c}{Facebook Perception-LM-1B} \\
Priority Signal & \multicolumn{2}{c}{$\bm{\mathrm{p}}^{\scriptscriptstyle\mathrm{VLM}} \cdot |\delta|$} \\
Priority Mode & \multicolumn{2}{c}{Hard Threshold ($>$ 0.5)} \\
Trajectory Length ($L$) & \multicolumn{2}{c}{32 Frames} \\
Importance Sampling & \multicolumn{2}{c}{False} \\
Prioritized Sampling Ratio ($\lambda_0$ and $\lambda_\mathrm{max}$) & \multicolumn{2}{c}{Annealed $0.0 \to 0.5$} \\
Annealing Schedule ($T_{\mathrm{schedule}}$) & \multicolumn{2}{c}{Linear over 500k steps} \\
\bottomrule
\end{tabular}
\end{table}

\newpage

\section{Ablations}
\label{app:ablations}

\subsection{Prioritization Variants}
\label{app:priority-variants}

To compare the two factors in \Eqref{eq:vlm_td_priority}, we evaluate uniform replay, TD-only priority, VLM-only priority, and their product on \texttt{DoorKey-16x16}/DQN and \texttt{Scene-5}/SAC. All VLM-bearing arms use the binarized instantiation with $\tau=0.5$. Each cell contains 5 seeds.

\begin{table}[H]
\centering
\caption{Prioritization variants on \texttt{DoorKey-16x16}/DQN and \texttt{Scene-5}/SAC (5 seeds per cell). Entries report ASR at 1M steps, followed in parentheses by the steps required to reach the best ASR attained by \fuer and \fpert, respectively. ``Never'' means that the target was not reached within 1M steps.}
\label{tab:priority-variants}
\begin{tabular}{lcc}
\toprule
\textbf{Priority} & \textbf{DoorKey-16x16} & \textbf{Scene-5} \\
\midrule
\fuer & 0.20 (920K, never) & 0.44 (900K, never) \\
\fpert (TD-only) & 0.76 (330K, 850K) & 0.62 (570K, 970K) \\
VLM-only & 1.00 (310K, 410K) & 0.88 (470K, 670K) \\
$\bm{\mathrm{p}}^{\scriptscriptstyle\mathrm{VLM}}|\delta|$ & 1.00 (310K, 410K) & 0.94 (320K, 440K) \\
\bottomrule
\end{tabular}
\end{table}

The VLM-only priority exceeds both baselines on both tasks. Multiplying it by $|\delta|$ leaves the DoorKey result unchanged and improves Scene-5 from 0.88 to 0.94 ASR, while reducing the steps to \fpert's best performance from 670K to 440K. We therefore use the product in the main experiments.

\subsection{Threshold Selection and the Threshold-Free Variant}
\label{app:threshold-ablation}

We vary the threshold in the binarized instantiation from \Appref{app:impl-details} and compare it with direct use of $r^{\scriptscriptstyle\mathrm{VLM}}$ as the priority. We analyze an offline corpus of 2,000 32-frame MiniGrid clips collected from policies at six stages of training, from random to fully trained. Strict success records whether the goal predicate becomes true within the clip. Table~\ref{tab:threshold-offline} reports the raw-score distribution and balanced accuracy as $\tau$ varies.

\begin{table}[H]
\centering
\caption{Offline threshold selection on the corpus. Bin share is the percentage of scores in each bin, indexed by its lower edge.}
\label{tab:threshold-offline}
\setlength{\tabcolsep}{3pt}
\begin{tabular}{@{}l*{9}{c}@{}}
\toprule
\textbf{Edge $\tau$} & 0.25 & 0.30 & 0.35 & 0.40 & 0.45 & 0.50 & 0.55 & 0.60 & 0.65 \\
\textbf{Bin share (\%)} & 1.45 & 13.95 & 10.30 & 33.05 & 11.30 & 10.50 & 16.00 & 3.20 & 0.25 \\
\textbf{Balanced accuracy} & 0.500 & 0.509 & 0.597 & 0.662 & 0.871 & 0.938 & 0.927 & 0.583 & 0.506 \\
\bottomrule
\end{tabular}
\end{table}

The raw-score distribution has modes near 0.40 and 0.55, and balanced accuracy peaks at the default threshold $\tau=0.5$. We next test whether this offline sensitivity predicts downstream performance and whether thresholding is necessary. These VLM-only ablations use \texttt{DoorKey-16x16}/DQN; the thresholded arms vary $\tau$ in \Eqref{eq:token_mass_ratio}, while the threshold-free arm uses $r^{\scriptscriptstyle\mathrm{VLM}}$ directly.

\begin{table}[H]
\centering
\caption{Downstream VLM-only threshold ablation on \texttt{DoorKey-16x16}/DQN (3 seeds per configuration). The threshold-free row uses the raw score $r^{\scriptscriptstyle\mathrm{VLM}}$ directly as the VLM-only priority. ASR is measured at 1M steps; steps to 0.8 are averaged over solved runs. Uncertainty is SEM over seeds.}
\label{tab:threshold-downstream}
\begin{tabular}{lcc}
\toprule
\textbf{Score form} & \textbf{ASR@1M} & \textbf{Steps to 0.8} \\
\midrule
$\tau=0.40$ & $0.53 \pm 0.27$ & $755\mathrm{K} \pm 125\mathrm{K}$ (2/3) \\
$\tau=0.45$ & $1.00 \pm 0.00$ & $547\mathrm{K} \pm 135\mathrm{K}$ \\
$\tau=0.50$ & $1.00 \pm 0.00$ & $357\mathrm{K} \pm 32\mathrm{K}$ \\
Threshold-free ($r^{\scriptscriptstyle\mathrm{VLM}}$) & $1.00 \pm 0.00$ & $537\mathrm{K} \pm 143\mathrm{K}$ \\
\bottomrule
\end{tabular}
\end{table}

The thresholds with the highest offline balanced accuracy both reach 1.00 ASR, while $\tau=0.40$ degrades performance. The threshold-free variant also reaches 1.00 ASR, but reaches 0.8 success at $537\mathrm{K} \pm 143\mathrm{K}$ steps, compared with $357\mathrm{K} \pm 32\mathrm{K}$ for the default threshold $\tau=0.5$. Thus offline calibration and direct use of the score provide two alternatives to selecting the threshold.

\subsection{Mixing Schedule}
\label{app:mix-schedule-ablation}
\begin{table}[H]
\centering
\footnotesize
\setlength{\tabcolsep}{1.5pt}
\caption{Ablation of mixing schedule ($\lambda_{\max}$) on \texttt{DoorKey-16x16} with VLM-only priority. \textbf{Performance} denotes the highest average success rate ($M=5$ seeds). \textbf{Sample efficiency} tracks the steps required to reach the \textit{baseline's} best performance.}
\label{tab:ablation-schedule}
\begin{tabular}{c c c c}
\toprule
& & \multicolumn{2}{c}{\textit{Improvement}} \\
\cmidrule(lr){3-4}
\textbf{$\lambda_{\max}$} & \textbf{Base.} & \textbf{Perf.} ($\uparrow$) & \textbf{Sample Eff.} ($\downarrow$) \\
\midrule
\multirow{2}{*}{\shortstack{Pure priority\\from start}}
 & \fuer & \res{N/A}{0.00}{0.24} & \res{N/A}{1000K}{916K} \\
 & \fpert & \res{N/A}{0.00}{0.80} & \res{N/A}{1000K}{592K} \\
\midrule
\multirow{2}{*}{0.25}
 & \fuer & \res{{+233.3\%}}{0.80}{0.24} & \res{{+40.6\%}}{544K}{916K} \\
 & \fpert & \res{\zerov{+0.0\%}}{0.80}{0.80} & \res{-0.3\%}{594K}{592K} \\
\midrule
\multirow{2}{*}{0.50}
 & \fuer & \res{\best{+316.7\%}}{1.00}{0.24} & \res{\best{+62.9\%}}{340K}{916K} \\
 & \fpert & \res{\best{+25.0\%}}{1.00}{0.80} & \res{\best{+35.5\%}}{382K}{592K} \\
\midrule
\multirow{2}{*}{0.75}
 & \fuer & \res{{+266.7\%}}{0.88}{0.24} & \res{{+57.0\%}}{394K}{916K} \\
 & \fpert & \res{{+10.0\%}}{0.88}{0.80} & \res{{+17.2\%}}{490K}{592K} \\
\midrule
\multirow{2}{*}{1.00}
 & \fuer & \res{\best{+316.7\%}}{1.00}{0.24} & \res{\best{+69.0\%}}{284K}{916K} \\
 & \fpert & \res{\best{+25.0\%}}{1.00}{0.80} & \res{\best{+40.2\%}}{354K}{592K} \\
\bottomrule
\end{tabular}
\end{table}

\begin{figure}[H]
    \centering
    \includegraphics[width=1\linewidth]{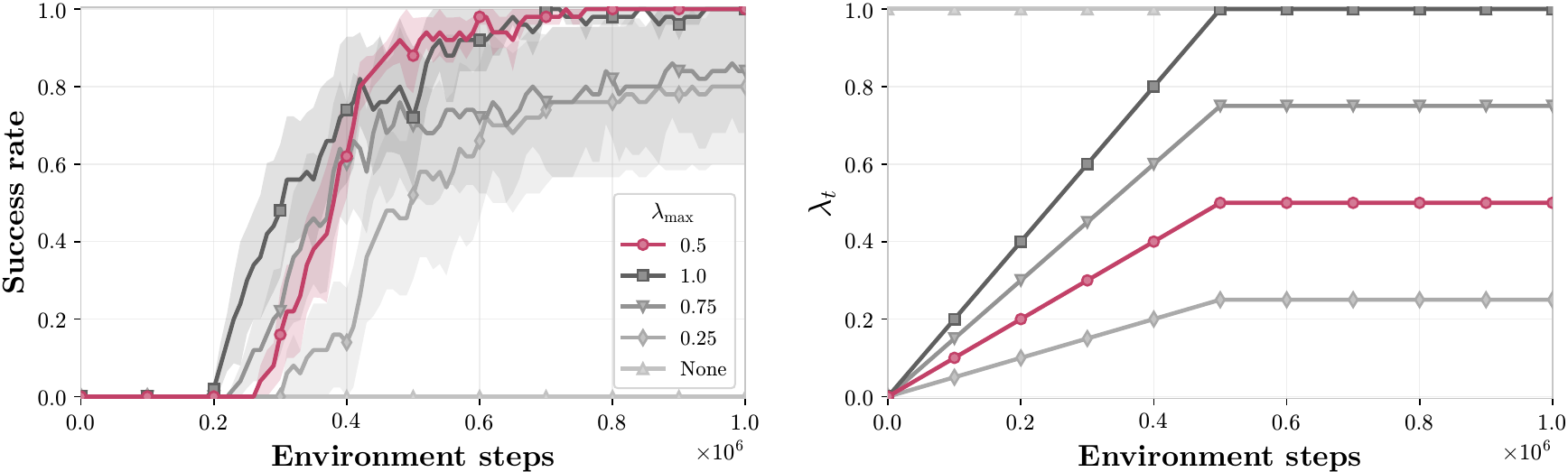}
    \caption{\texttt{MiniGrid/DoorKey-16x16}: $\lambda_\mathrm{max}$ ablation (5 seeds). ``Pure priority from start'' disables the warm-up and sets $\lambda_t=1$ throughout. Left: performance for different $\lambda_\mathrm{max}$ values. Right: $\lambda_\mathrm{max}$ value evolution.}
    \label{fig:mixing-schedule}
\end{figure}

To understand the effect of the final mixing coefficient $\lambda_{\max}$, we conduct an ablation on the \texttt{MiniGrid/DoorKey-16x16} task (Fig.\ref{fig:mixing-schedule}). We fix $T_{\mathrm{schedule}}=5\cdot10^5$, ensuring that $\lambda_t$ is annealed linearly from $0$ (corresponding to purely uniform sampling) to $\lambda_{\max}$ over the first half of training (right panel).
We sweep $\lambda_{\max}\in\{0.25,0.5,0.75,1.0\}$ and include a \textbf{pure-priority-from-start} baseline, which disables the schedule and sets $\lambda_t=1$ from the first update.
We observe that larger $\lambda_{\max}$ values reach $50\%$ success marginally earlier than smaller values. Both $\lambda_{\max}=0.5$ and the scheduled $\lambda_{\max}=1.0$ arm reach a final ASR of $1.00$, while $\lambda_{\max}=0.75$ reaches $0.88$.
In contrast, pure priority from the first update fails to solve the task, with success rates remaining near zero. This suggests that mixing uniform samples into replay during early training is important for effective learning in this setting.
We fix $\lambda_{\max}=0.5$ for all main experiments to preserve uniform coverage, without further per-task tuning.

To separate score granularity from the mixture, we additionally use raw-score priorities. VLM-only reaches ASR $1.00\pm0.00$ with the default schedule and $0.07\pm0.07$ with no mixture, while the raw-score product $r^{\scriptscriptstyle\mathrm{VLM}}|\delta|$ with no mixture reaches $0.00\pm0.00$ (three seeds each). Together with Table~\ref{tab:ablation-schedule}'s hard-score results---$1.00$ at $\lambda_{\max}=0.5$ and $0.00$ for pure priority from the start (five seeds)---changing score granularity does not rescue fully prioritized sampling. Here, no mixture sets $\lambda_t=1$ throughout; the prioritized sampler retains only its $10^{-6}$ numerical floor and draws no uniform samples.

\subsection{Realized Sampling Frequencies}
\label{app:sampling-frequencies}

At 250K steps in \texttt{Scene-5}/SAC, we count replay draws for each transition class and divide by the number expected under uniform replay. Table~\ref{tab:sampling-frequencies} reports the mean across five seeds using signal $\bm{\mathrm{p}}^{\scriptscriptstyle\mathrm{VLM}}|\delta|$. The top-level frequencies have a class-size-weighted mean of $1.0$. VLM-negative transitions constitute $81.1\%$ of the buffer and are sampled at $0.91\times$ the uniform rate.

\begin{table}[H]
\centering
\caption{Realized replay frequencies at the 250K buffer snapshot in \texttt{Scene-5}/SAC (five seeds). The successful-episode subset is contained within the VLM-positive row. A ratio of $1.0$ is uniform replay, and $0.5$ is the instantaneous mixture floor. Unscored transitions are newest, so their cumulative ratio may be lower because they have been present for fewer sampling steps.}
\label{tab:sampling-frequencies}
\begin{tabular}{lcc}
\toprule
\textbf{Transition class} & \textbf{Buffer share} & \textbf{Frequency / uniform} \\
\midrule
VLM-positive & $18.8\%$ & $1.41\times$ \\
Successful-episode subset & $1.4\%$ & $2.17\times$ \\
VLM-negative & $81.1\%$ & $0.91\times$ \\
No VLM score yet & $0.1\%$ & $0.44\times$ \\
\bottomrule
\end{tabular}
\end{table}

\subsection{VLM Size}
\label{app:vlm-size-ablation}

\begin{table}[H]
\centering
\caption{Performance comparison of Perception-LM~\citep{cho2025PerceptionLM} models. Values in parentheses denote the relative percentage change compared to 1B. We use an Nvidia RTX 4090 GPU, and run 100 batches of 32 frames.}
\label{tab:performance_comparison}
\resizebox{0.6\linewidth}{!}
    {
        \begin{tabular}{lcccc}
        \toprule
        \textbf{Model} & \textbf{Load (GiB)} & \textbf{Peak (GiB)} & \textbf{Time (s)} & \textbf{FPS} \\
        \midrule
        1B & 2.86 & 3.77 & 0.46 & 69.27 \\
        3B & 6.56 \scriptsize{(+130\%)} & 8.16 \scriptsize{(+116\%)} & 0.77 \scriptsize{(+66\%)} & 41.75 \scriptsize{(-40\%)} \\
        8B & 18.25 \scriptsize{(+539\%)} & 20.34 \scriptsize{(+439\%)} & 2.15 \scriptsize{(+366\%)} & 14.88 \scriptsize{(-79\%)} \\
        \bottomrule
        \end{tabular}
    }
\label{tab:vlm_size_performance_comparison}
\end{table}

We investigate how scaling the VLM affects both inference overhead and downstream RL performance. Table~\ref{tab:vlm_size_performance_comparison} quantifies the resource requirements of Perception-LM~\citep{cho2025PerceptionLM} variants for clip scoring (Nvidia RTX 4090 GPU; 100 batches of 32-frame clips).
We observe that increasing the VLM size leads to a substantial increase in memory footprint and a corresponding reduction in throughput. Relative to the 1B model, the 3B variant increases peak memory by $+116\%$ and reduces FPS by $40\%$; the 8B variant increases peak memory by $+439\%$ and reduces FPS by $79\%$.

\begin{figure}[H]
    \centering
    \includegraphics[width=0.5\linewidth]{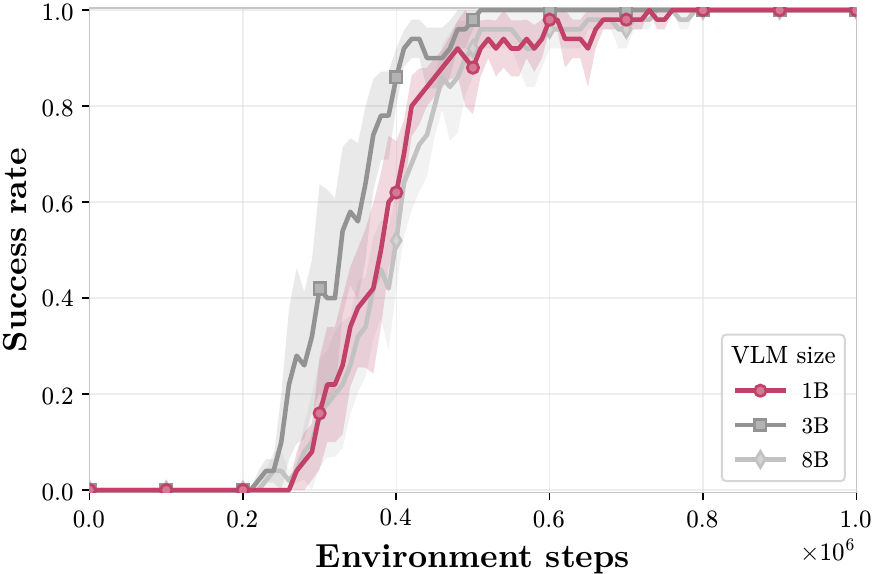}
    \caption{\texttt{MiniGrid/DoorKey-16x16}: VLM size ablation (5 seeds)}
    \label{fig:vlm_size}
\end{figure}

To assess the impact of VLM size on downstream performance, we compare the 1B, 3B, and 8B models on \texttt{MiniGrid/DoorKey-16x16} (Fig.~\ref{fig:vlm_size}). Notably, despite the increased inference cost of larger models, we do not observe consistent improvements in RL performance relative to the 1B configuration.
These results suggest that in this setting, the clip-scoring signal saturates once the VLM is sufficiently reliable at separating goal-consistent from goal-inconsistent segments. Given this trade-off, we select the 1B Perception-LM as the default backbone for all main experiments.

\subsection{VLM Family Dependence}
\label{app:vlm-family-dependence}

We score the offline corpus with three VLM families using the same prompt. Each threshold is selected on one half of the corpus and evaluated on the disjoint half. Table~\ref{tab:vlm-family-offline} reports held-out strict-success classification and full-corpus agreement with Perception-LM-1B.

\begin{table}[H]
\centering
\caption{VLM-family diagnostics on the offline corpus. Balanced accuracy and AUC use the held-out half; agreement and raw-score Spearman $\rho$ use the full corpus relative to Perception-LM-1B.}
\label{tab:vlm-family-offline}
\begin{tabular}{lccccc}
\toprule
\textbf{Model} & $\bm{\tau}$ & \textbf{Bal. acc.} & \textbf{AUC} & \textbf{Agreement} & $\bm{\rho}$ \\
\midrule
Perception-LM-1B & 0.50 & 0.94 & 0.98 & 100\% & 1.00 \\
Gemma-4-E4B & 0.12 & 0.88 & 0.92 & 86\% & 0.70 \\
Qwen3-VL-2B & 0.89 & 0.85 & 0.89 & 81\% & 0.73 \\
\bottomrule
\end{tabular}
\end{table}

The raw scores rank clips similarly across families, but their scales differ. We therefore compare offline threshold calibration with direct use of the raw score on \texttt{DoorKey-16x16}/DQN.

\begin{table}[H]
\centering
\caption{VLM-only downstream comparison on \texttt{DoorKey-16x16}/DQN. The Perception-LM row contains 3 seeds; both Gemma rows contain 5. ASR is measured at 1M steps and uncertainty is SEM over seeds.}
\label{tab:vlm-family-downstream}
\begin{tabular}{lllc}
\toprule
\textbf{Model} & $\bm{f_{\scriptscriptstyle\mathrm{VLM}}}$ & \textbf{Calibration} & \textbf{ASR@1M} \\
\midrule
Perception-LM-1B & Raw score & None & $1.00\pm0.00$ \\
Gemma-4-E4B & Raw score & None & $0.80\pm0.15$ \\
Gemma-4-E4B & $\mathbf{1}[r^{\scriptscriptstyle\mathrm{VLM}}>0.12]$ & Offline threshold & $1.00\pm0.00$ \\
\bottomrule
\end{tabular}
\end{table}

For Gemma, the calibrated threshold solves all five seeds; using the raw score requires no calibration and reaches ASR $0.80\pm0.15$. Thus a new VLM/prompt combination requires either a labeled calibration corpus for the hard-threshold variant or acceptance of the raw-score variant's observed performance trade-off.

\subsection{Scorer Reliability}
\label{app:scorer-reliability}

We evaluate the default Perception-LM-1B score against two simulator-derived labels on the offline corpus. Strict success records whether the goal predicate becomes true within the clip; milestone acquisition records key pickup, door opening, or the goal being reached.

\begin{table}[H]
\centering
\caption{Reliability of the default binarized scorer at $\tau=0.5$ on the offline corpus.}
\label{tab:scorer-reliability}
\begin{tabular}{lcccc}
\toprule
\textbf{Ground-truth definition} & \textbf{Precision} & \textbf{Recall} & \textbf{FPR} & \textbf{FNR} \\
\midrule
Strict success & 0.69 & 0.99 & 0.12 & 0.01 \\
Milestone acquisition & 0.92 & 0.62 & 0.04 & 0.38 \\
\bottomrule
\end{tabular}
\end{table}

Under strict success, the scorer produces 185 false positives and 3 false negatives. Of the false positives, 137 contain milestone acquisition and 29 more show positive net progress, leaving 19 clips with neither event ($0.95\%$ of the corpus and $3.2\%$ of positive labels).

\subsection{Robustness to Label Error}
\label{app:label-corruption}

We test controlled label error on \texttt{DoorKey-16x16}/DQN using VLM-only priority and the binarized scorer. After thresholding, each positive clip label is demoted independently with probability $\kappa$, and the same number of negative clip labels are promoted uniformly without replacement. Unmatched demotions are carried to later scoring batches. The corruption changes training-time replay priorities only; rewards, bootstrap targets, and evaluation are unchanged.

\begin{table}[H]
\centering
\caption{Controlled clip-label corruption on \texttt{DoorKey-16x16}/DQN with VLM-only priority (five seeds per configuration; mean ASR at 1M steps $\pm$ SEM). The deficit is the worst seed's terminal uncompensated demotions as a fraction of the cumulative positive-label count.}
\label{tab:label-corruption}
\begin{tabular}{lccccc}
\toprule
$\bm{\kappa}$ & 0.00 & 0.10 & 0.25 & 0.50 & 0.75 \\
\midrule
\textbf{ASR@1M} & $1.00 \pm 0.00$ & $0.96 \pm 0.04$ & $1.00 \pm 0.00$ & $0.40 \pm 0.24$ & $0.44 \pm 0.20$ \\
\textbf{vs. UER (0.20)} & $+0.80$ & $+0.76$ & $+0.80$ & $+0.20$ & $+0.24$ \\
\textbf{Worst deficit} & $0.0\%$ & $0.0\%$ & $0.1\%$ & $4.0\%$ & $21.4\%$ \\
\bottomrule
\end{tabular}
\end{table}

Table~\ref{tab:label-corruption} reports five seeds per corruption level, with $\kappa=0$ reusing the uncorrupted VLM-only arm. ASR remains $0.96$--$1.00$ through $\kappa=0.25$, well above the measured $3.2\%$ no-event label rate from \Appref{app:scorer-reliability}, and falls to $0.40$--$0.44$ at the two higher levels. Paired substitution preserves the cumulative positive-label count within $0.1\%$ for $\kappa\leq0.25$; the higher-corruption points also reduce the prioritized set and therefore do not isolate precision alone.

\subsection{Prompt Sensitivity}
\label{app:prompt-sensitivity}

We re-score the offline corpus with the following four variants of the \texttt{MiniGrid/DoorKey} prompt.
\begin{enumerate}[leftmargin=*, topsep=2pt, itemsep=2pt]
    \item \textbf{Paraphrase A.}
    \emph{``Does this video clip visibly show the agent successfully completing a task goal or subgoal? Answer No when no clear success is visible, and do not infer hidden progress. Output exactly Answer: Yes or Answer: No.''}

    \item \textbf{Paraphrase B.}
    \emph{``Is there clear visual evidence anywhere in this clip that the agent achieves an intended goal state? If the evidence is absent or ambiguous, answer No. Output exactly Answer: Yes or Answer: No.''}

    \item \textbf{Terse.}
    \emph{``Is a task goal or subgoal visibly achieved in this clip? Output exactly Answer: Yes or Answer: No.''}

    \item \textbf{Underspecified.}
    \emph{``Does anything important happen in this clip? Output exactly Answer: Yes or Answer: No.''}
\end{enumerate}
Table~\ref{tab:prompt-sensitivity} reports fixed-threshold label agreement and raw-score rank correlation. The two semantic paraphrases preserve $90.9$--$95.3\%$ of labels. The terse prompt labels every clip positive at $\tau=0.5$, while the underspecified prompt labels every clip negative; their raw-score rankings remain correlated with the other prompts, and all five scores retain strict-success ROC AUC $0.955$--$0.989$.

\begin{table}[H]
\centering
\caption{Prompt sensitivity on the offline corpus. Lower-triangular entries are fixed-$0.5$ label agreement / raw-score Spearman correlation; the last column is strict-success ROC AUC.}
\label{tab:prompt-sensitivity}
\setlength{\tabcolsep}{5pt}
\begin{tabular}{lcccccc}
\toprule
\textbf{Prompt} & \textbf{Base} & \textbf{Paraph. A} & \textbf{Paraph. B} & \textbf{Terse} & \textbf{Under.} & \textbf{AUC} \\
\midrule
Base & -- & & & & & 0.989 \\
Paraphrase A & 0.909 / 0.892 & -- & & & & 0.989 \\
Paraphrase B & 0.945 / 0.956 & 0.953 / 0.911 & -- & & & 0.987 \\
Terse & 0.300 / 0.931 & 0.210 / 0.857 & 0.254 / 0.933 & -- & & 0.983 \\
Underspecified & 0.701 / 0.881 & 0.791 / 0.813 & 0.747 / 0.909 & 0.000 / 0.858 & -- & 0.955 \\
\bottomrule
\end{tabular}
\end{table}

The most divergent semantic paraphrase, Paraphrase A, reaches ASR $0.84\pm0.16$ on \texttt{DoorKey-16x16} with VLM-only raw-score priority (five seeds).

\subsection{LavaCrossing Analysis}
\label{app:lava-analysis}

\begin{figure}[H]
    \centering
    \includegraphics[width=0.96\linewidth]{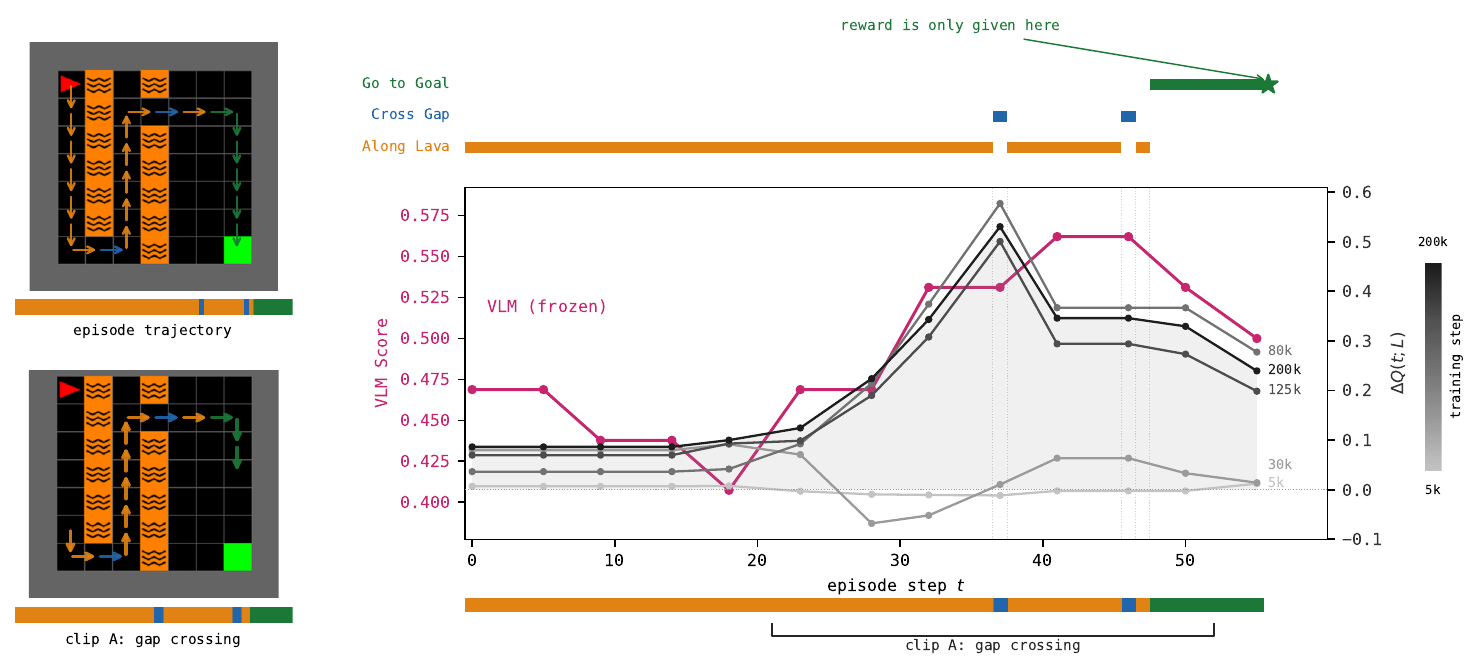}
    \caption{\textbf{Fig.~\ref{fig:vlm_vs_Q} protocol on \texttt{LavaCrossing}.} On a successful episode, the frozen VLM score anticipates the learned value signal around the gap crossing.}
    \label{fig:lava-analysis}
\end{figure}

We apply the analysis from Fig.~\ref{fig:vlm_vs_Q} to \texttt{MiniGrid/LavaCrossing-S9N2}, where stepping into lava terminates the episode with zero reward. On a successful episode, the frozen VLM score crosses the threshold before the agent commits through the lava gaps, while the early critics remain flat ($|\Delta Q|\leq 0.03$); only the converged critic later develops the corresponding $\Delta Q$ surge (Fig.~\ref{fig:lava-analysis}).

On a fatal episode, a large negative state-value difference does not appear. At the pre-terminal state, $Q(s,a_{\mathrm{fatal}})$ reaches $0.109$ at 55K training steps and falls to $-0.007$ at 200K, while $V(s)=\max_a Q(s,a)$ reaches $0.760$. The learned hazard is action-conditional: stepping away remains viable, so a state-value curve cannot display the fatal action's value; the replay signal remains action-conditioned through its per-transition $|\delta|$.

\subsection{VLM Prior}
\label{sec:vlm_prior}
Motivated by the ``Modified Game'' paradigm of \citet{dubey2018investigating}, we probe whether \fmethod leverages the VLM's pre-trained visual semantics to improve performance.
To isolate the effect of visual semantics, we modify only the rendered frames provided to the VLM for scoring, while leaving both the underlying MDP and the agent's observations unchanged.
If \fmethod relies on semantic cues such as identifying keys, doors, or goal-relevant interactions, we expect its performance gains to diminish when these cues are distorted or removed.

\begin{figure}[H]
    \centering

    \begin{tabular}{@{}c@{\hspace{0.04\textwidth}}c@{\hspace{0.04\textwidth}}c@{}}

        \begin{minipage}[t]{0.26\textwidth}
            \centering
            \includegraphics[width=\linewidth]{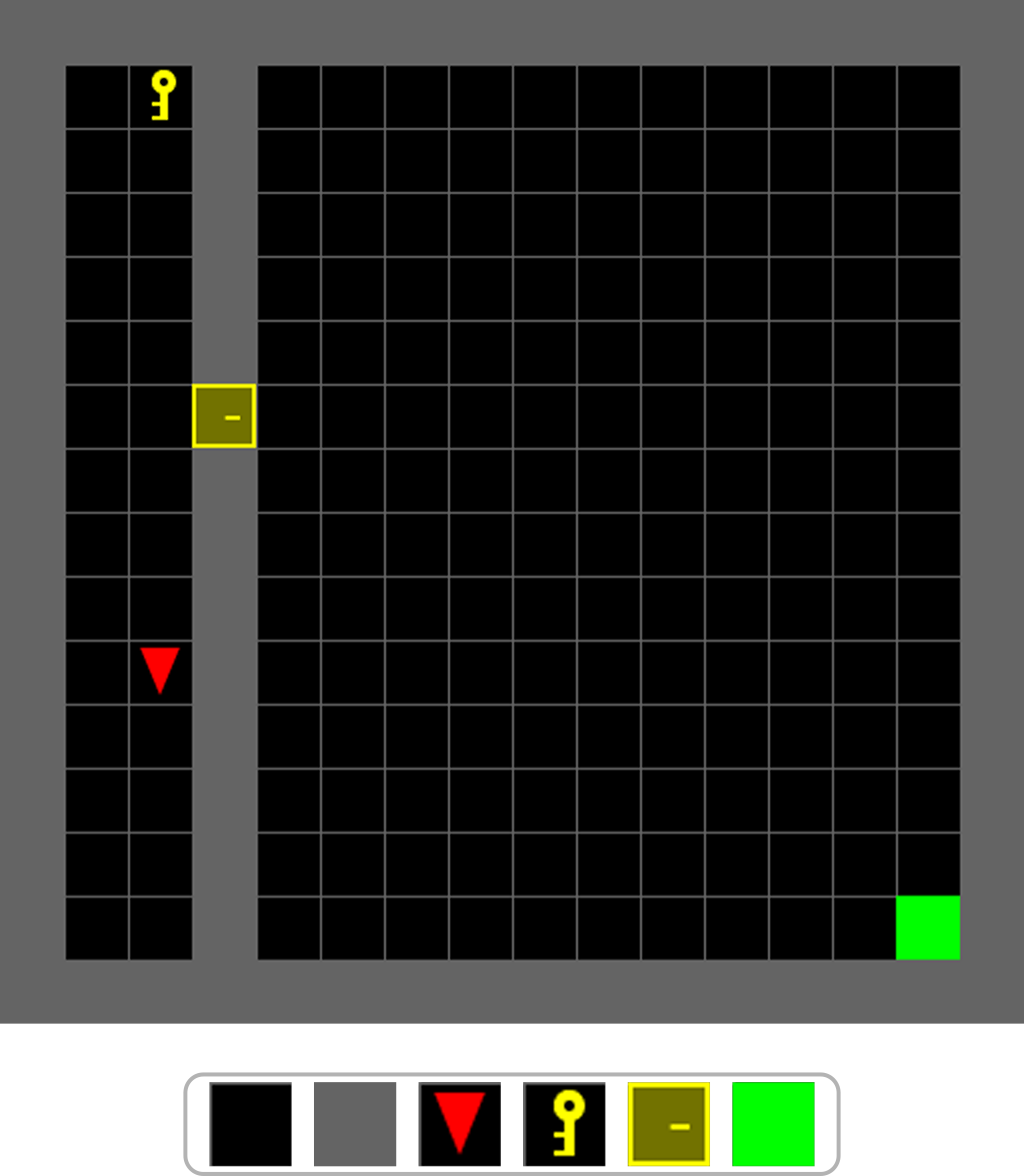}

            \vspace{0.4em}
            \parbox[t][3.2em][t]{\linewidth}{\centering\small
                (a) Standard rendering
            }
        \end{minipage}
        &
        \begin{minipage}[t]{0.26\textwidth}
            \centering
            \includegraphics[width=\linewidth]{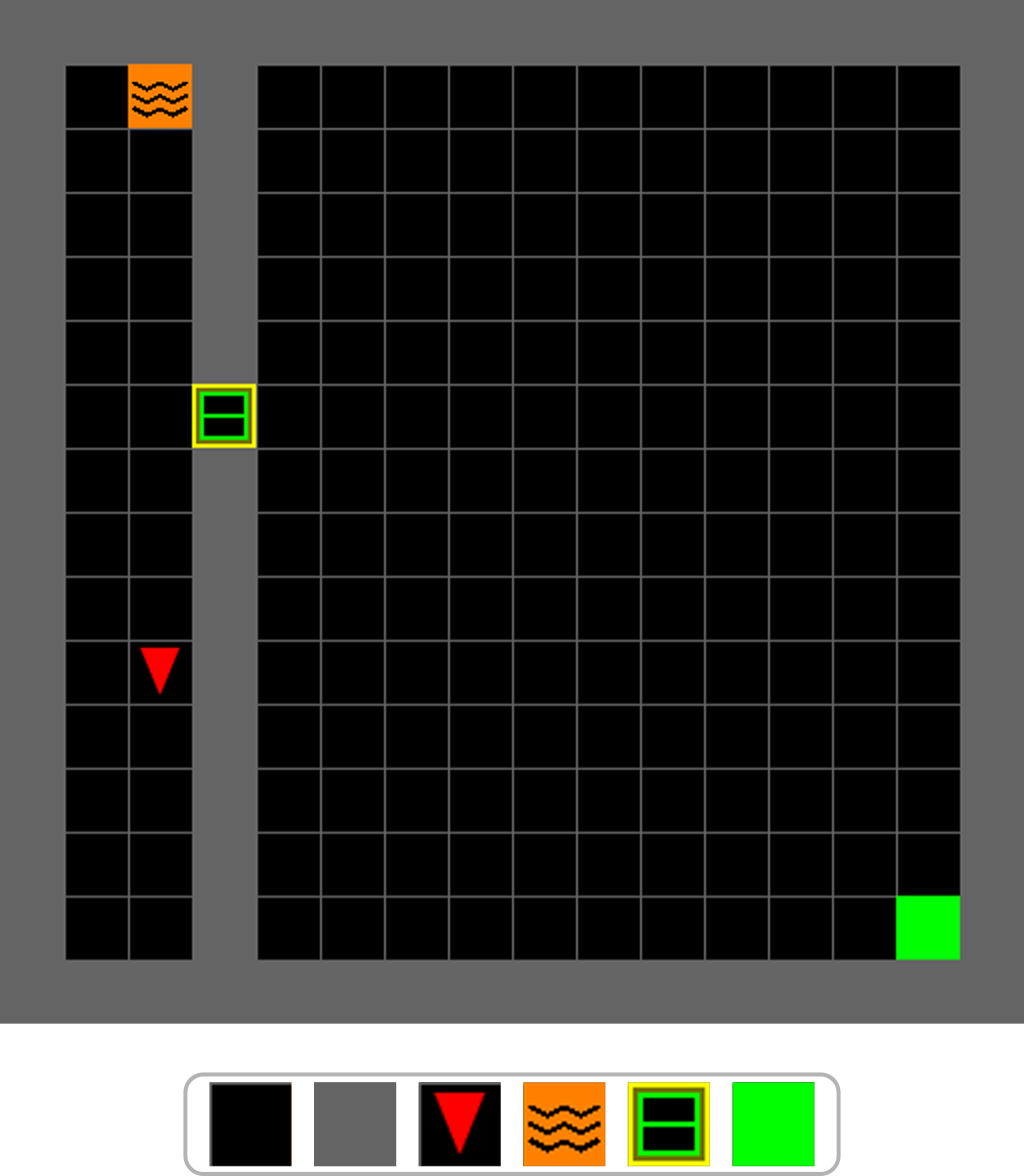}

            \vspace{0.4em}
            \parbox[t][3.2em][t]{\linewidth}{\centering\small
                (b) Semantically misleading sprites
            }
        \end{minipage}
        &
        \begin{minipage}[t]{0.26\textwidth}
            \centering
            \includegraphics[width=\linewidth]{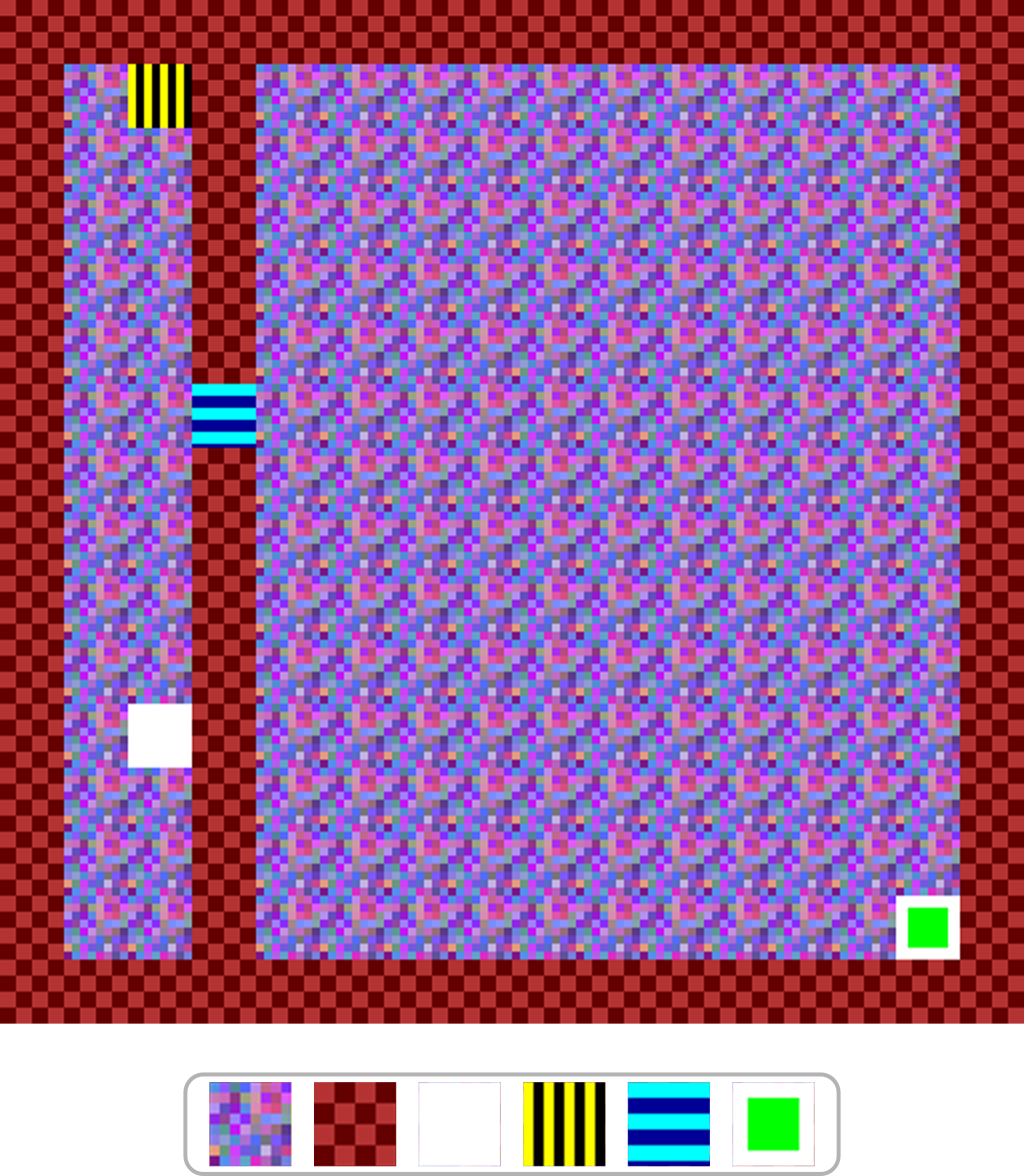}

            \vspace{0.4em}
            \parbox[t][3.2em][t]{\linewidth}{\centering\small
                (c) Abstract patterns that remove naturalistic cues
            }
        \end{minipage}

    \end{tabular}

    \caption{Samples of the modified visuals. We modify only the frames passed to the VLM for scoring (agent observations and environment dynamics unchanged).}
    \label{fig:vlm-prior-envs}
\end{figure}

To test this, we introduce two renderer perturbations (Fig.~\ref{fig:vlm-prior-envs}). First, \texttt{Sprite Swap} replaces object sprites with semantically conflicting alternatives, such as rendering keys as lava or doors as boxes, thereby introducing misleading visual priors. Second, \texttt{Texture} replaces all object appearances with abstract high-contrast patterns, removing naturalistic semantics entirely.

Empirically, we observe that the unmodified setting achieves near-perfect success and converges the fastest (Fig.~\ref{fig:vlm_prior_ablation}).
In contrast, both \texttt{Sprite Swap} and \texttt{Texture} slow learning and reduce final success rates, with \texttt{Texture} also leading to the largest variance across seeds.
Because the control problem and agent inputs remain fixed, this degradation suggests that the VLM produces less informative priorities when visual evidence for goal-relevant events is either misleading or lacks semantic content.

\subsection{Computational Overhead and Scoring Lag}
\label{app:throughput-ablation}

\begin{table}[H]
\centering
\caption{\textbf{Training throughput} (steps/second) comparison on \texttt{MiniGrid/DoorKey-16x16}. Higher is better.}
\resizebox{0.5\linewidth}{!}
    {
    \begin{tabular}{l c c c}
        \toprule
        \textbf{Hardware} & \textbf{\fpert} & \textbf{\fmethod} & \textbf{Rel. Speed} \\
        \midrule
        NVIDIA A100 & 111 & 97 & $87\%$ \\
        NVIDIA A40 & 92 & 81 & $88\%$ \\
        NVIDIA A4000 & 76 & 67 & $88\%$ \\
        \bottomrule
    \end{tabular}
    }
\label{tab:throughput}
\end{table}

How much does \fmethod actually slow down training in practice? To answer this, we measure throughput (steps per second) on the \texttt{MiniGrid/DoorKey-16x16} task using DQN, comparing \fpert and \fmethod across three dual-GPU setups (NVIDIA A100, A40, and A4000). In each case, the RL learner and VLM are placed on separate devices. Notably, in this distributed configuration, the main bottleneck is inter-process communication (IPC) and data transfer, rather than competition for computational resources.

The results, summarized in Table~\ref{tab:throughput}, reveal a consistent and modest throughput reduction of about $12\%$ across all hardware types. This suggests that, even when VLM inference is separated from the RL learner, the method maintains efficient scaling. In other words, the additional overhead introduced by \fmethod remains limited in practical distributed settings.

\textbf{Scoring lag.}
Across 20 logged \texttt{DoorKey-16x16} runs, the scorer keeps pace with collection: a backlog-based estimate gives zero median delay, the queues are empty at $99.7\%$ of logged steps, the worst transient is $10.5$K steps, and no frame is dropped.

It is important to note that these measurements use a standard inference setup, without any aggressive optimizations such as TensorRT or quantization (e.g., INT8 or FP4). We anticipate that employing a dedicated serving stack, such as vLLM or TGI, would further close the speed gap between \fmethod and the baseline.

\section{Experiments}
\label{app:experiments}
In this section, we present the full results of the baselines on all environments.

\begin{table}[H]
\centering
\caption{
\textbf{Performance} denotes the mean per-seed best success rate ($M=5$ seeds).
\textbf{Sample efficiency} tracks the steps required to reach the \textit{baseline's} best performance; unreached targets are assigned the 1M-step horizon.
\textbf{Wall-clock saving} estimates the reduction in real training time, accounting for a $12\%$ inference overhead per step.
}
\label{tab:experiments-full}
\resizebox{\linewidth}{!}{%
\begin{tabular}{ll c c c c}
\toprule
 & & & \textbf{Performance} ($\uparrow$) & \textbf{Sample Efficiency} ($\downarrow$) & \textbf{Wall-Clock Saving} ($\uparrow$) \\
\textbf{Alg.} & \textbf{Task} & \textbf{Baseline} & \textit{Best ASR} & \textit{Steps to \textit{Base.} Best} & \textit{Time vs. Baseline} \\
\midrule
\multirow{6}{*}{SAC}
 & \multirow{2}{*}{\texttt{Scene-3}} & \fuer & \res{\zerov{+0.0\%}}{1.00}{1.00} & \res{\best{+50.7\%}}{224K}{454K} & \best{+44.7\%} \\
 &  & \fpert & \res{\zerov{+0.0\%}}{1.00}{1.00} & \res{\best{+18.8\%}}{224K}{276K} & \best{+9.1\%} \\
 \cmidrule{2-6}
 & \multirow{2}{*}{\texttt{Scene-4}} & \fuer & \res{\best{+4.2\%}}{1.00}{0.96} & \res{\best{+54.2\%}}{378K}{826K} & \best{+48.7\%} \\
 &  & \fpert & \res{\best{+4.2\%}}{1.00}{0.96} & \res{\best{+52.8\%}}{378K}{800K} & \best{+47.1\%} \\
 \cmidrule{2-6}
 & \multirow{2}{*}{\texttt{Scene-5}} & \fuer & \res{\best{+113.6\%}}{0.94}{0.44} & \res{\best{+60.0\%}}{366K}{914K} & \best{+55.2\%} \\
 &  & \fpert & \res{\best{+51.6\%}}{0.94}{0.62} & \res{\best{+34.7\%}}{494K}{756K} & \best{+26.8\%} \\
\midrule
\multirow{6}{*}{TD3}
 & \multirow{2}{*}{\texttt{Scene-3}} & \fuer & \res{\zerov{+0.0\%}}{1.00}{1.00} & \res{\best{+20.5\%}}{202K}{254K} & \best{+10.9\%} \\
 &  & \fpert & \res{\zerov{+0.0\%}}{1.00}{1.00} & \res{\best{+21.1\%}}{202K}{256K} & \best{+11.6\%} \\
 \cmidrule{2-6}
 & \multirow{2}{*}{\texttt{Scene-4}} & \fuer & \res{\best{+47.1\%}}{1.00}{0.68} & \res{\best{+70.8\%}}{212K}{726K} & \best{+67.3\%} \\
 &  & \fpert & \res{\zerov{+0.0\%}}{1.00}{1.00} & \res{\best{+12.4\%}}{410K}{468K} & \best{+1.9\%} \\
 \cmidrule{2-6}
 & \multirow{2}{*}{\texttt{Scene-5}} & \fuer & \res{\best{+200.0\%}}{0.84}{0.28} & \res{\best{+49.4\%}}{366K}{724K} & \best{+43.4\%} \\
 &  & \fpert & \res{\best{+90.9\%}}{0.84}{0.44} & \res{\best{+38.9\%}}{522K}{854K} & \best{+31.5\%} \\
\midrule
\multirow{6}{*}{DQN}
 & \multirow{2}{*}{\texttt{DoorKey-8x8}} & \fuer & \res{\zerov{+0.0\%}}{1.00}{1.00} & \res{\best{+14.3\%}}{144K}{168K} & \best{+4.0\%} \\
 &  & \fpert & \res{\zerov{+0.0\%}}{1.00}{1.00} & \res{\best{+30.8\%}}{144K}{208K} & \best{+22.5\%} \\
 \cmidrule{2-6}
 & \multirow{2}{*}{\texttt{DoorKey-12x12}} & \fuer & \res{\best{+66.7\%}}{1.00}{0.60} & \res{\best{+55.4\%}}{258K}{578K} & \best{+50.0\%} \\
 &  & \fpert & \res{\zerov{+0.0\%}}{1.00}{1.00} & \res{-9.9\%}{288K}{262K} & -23.1\% \\
 \cmidrule{2-6}
 & \multirow{2}{*}{\texttt{DoorKey-16x16}} & \fuer & \res{\best{+316.7\%}}{1.00}{0.24} & \res{\best{+64.2\%}}{328K}{916K} & \best{+59.9\%} \\
 &  & \fpert & \res{\best{+25.0\%}}{1.00}{0.80} & \res{\best{+40.2\%}}{354K}{592K} & \best{+33.0\%} \\
\midrule
\multirow{6}{*}{IQN}
 & \multirow{2}{*}{\texttt{DoorKey-8x8}} & \fuer & \res{\zerov{+0.0\%}}{1.00}{1.00} & \res{\best{+28.7\%}}{134K}{188K} & \best{+20.2\%} \\
 &  & \fpert & \res{\zerov{+0.0\%}}{1.00}{1.00} & \res{\best{+19.3\%}}{134K}{166K} & \best{+9.6\%} \\
 \cmidrule{2-6}
 & \multirow{2}{*}{\texttt{DoorKey-12x12}} & \fuer & \res{\best{+56.2\%}}{1.00}{0.64} & \res{\best{+57.8\%}}{296K}{702K} & \best{+52.8\%} \\
 &  & \fpert & \res{\best{+56.2\%}}{1.00}{0.64} & \res{\best{+56.0\%}}{296K}{672K} & \best{+50.7\%} \\
 \cmidrule{2-6}
 & \multirow{2}{*}{\texttt{DoorKey-16x16}} & \fuer & \res{\best{+250.0\%}}{0.84}{0.24} & \res{\best{+41.1\%}}{512K}{870K} & \best{+34.1\%} \\
 &  & \fpert & \res{\best{+425.0\%}}{0.84}{0.16} & \res{\best{+60.7\%}}{256K}{652K} & \best{+56.0\%} \\
\bottomrule
\end{tabular}
}
\end{table}

\begin{table}[ht]
\centering
\caption{\textbf{Performance} denotes the mean per-seed best success rate ($M=5$ seeds). \textbf{Sample efficiency} tracks the steps required to reach the \textit{baseline's} best performance. Both metrics are averaged across the aggregated algorithms and tasks.}
\label{tab:experiments-agg-algo-task}
\begin{tabular}{ll c c c}
\toprule
 & & & \textbf{Performance} ($\uparrow$) & \textbf{Sample Efficiency} ($\downarrow$) \\
\textbf{Env Type} & \textbf{Agg. Algorithms} & \textbf{Baseline} & \textit{Mean Best ASR} & \textit{Mean Steps to Base Peak} \\
\midrule
\multirow{2}{*}{\textbf{Scene}} & \multirow{2}{*}{\small{(SAC + TD3)}} & \fuer & \res{\best{+32.6\%}}{0.96}{0.73} & \res{\best{+55.2\%}}{291K}{650K} \\
 & & \fpert & \res{\best{+15.1\%}}{0.96}{0.84} & \res{\best{+34.6\%}}{372K}{568K} \\
\midrule
\multirow{2}{*}{\textbf{DoorKey}} & \multirow{2}{*}{\small{(DQN + IQN)}} & \fuer & \res{\best{+57.0\%}}{0.97}{0.62} & \res{\best{+51.1\%}}{279K}{570K} \\
 & & \fpert & \res{\best{+27.0\%}}{0.97}{0.77} & \res{\best{+42.3\%}}{245K}{425K} \\
\bottomrule
\end{tabular}
\end{table}

\begin{figure}[H]
    \centering
    \includegraphics[width=\linewidth]{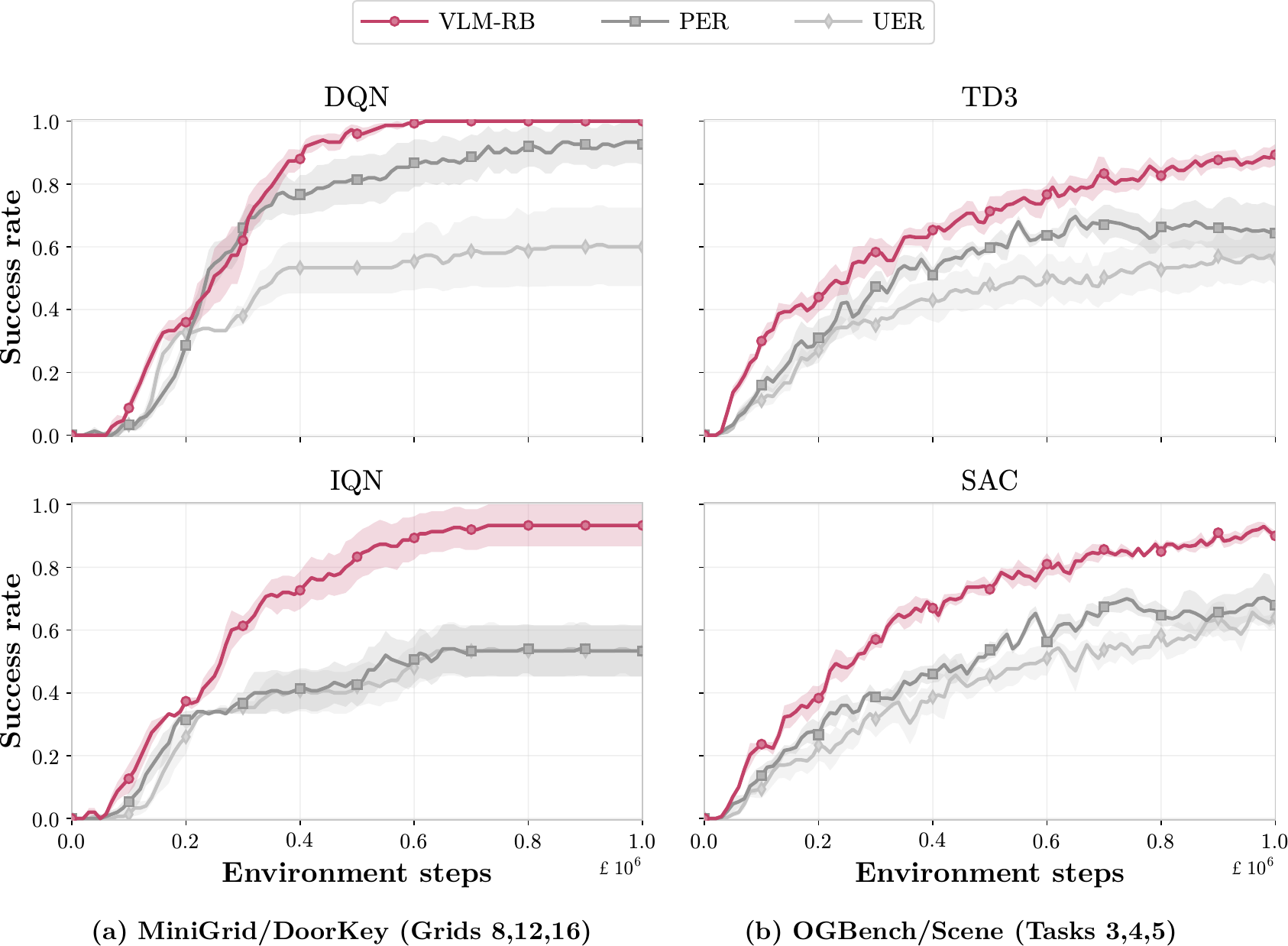}
    \caption{
    \fmethod \textbf{consistently outperforms baselines across continuous and discrete tasks.} The plots show aggregated success rates for four algorithms (DQN, IQN, SAC, TD3) on MiniGrid and OGBench domains. Shaded regions indicate standard error across seeds.
    }
    \label{fig:training-curves}
\end{figure}

\end{document}

%% file: custom_commands.tex
\usepackage{bm}
\usepackage{xspace}
\usepackage{tcolorbox}

\def\Eqref#1{Equation~\ref{#1}}
\def\gB{{\mathcal{B}}}
\def\gC{{\mathcal{C}}}
\def\gD{{\mathcal{D}}}
\def\gQ{{\mathcal{Q}}}

\definecolor{flare4}{RGB}{194,65,104}
\definecolor{blue}{RGB}{28,106,137}

\newcommand{\flare}[2]{\textbf{\textcolor{flare#1}{#2}}}
\newcommand{\method}{\textsc{VLM-RB}}
\newcommand{\uer}{\textsc{UER}}
\newcommand{\pert}{\textsc{PER}}
\newcommand{\fmethod}{\flare{4}{\method}\xspace}
\newcommand{\fuer}{\textbf{\textcolor{black!60}{\uer}}\xspace}
\newcommand{\fpert}{\textbf{\textcolor{black!80}{\pert}}\xspace}

\usepackage{algorithm}
\usepackage[endLComment=,italicComments=false]{algpseudocodex}

\usepackage{enumitem}

\def\Appref#1{Appendix~\ref{#1}}

\tcbuselibrary{breakable,skins}

\newtcolorbox{promptbox}[1][]{%
  breakable,
  enhanced,
  title={#1},
  colframe=black!50,
  colback=white,
  coltitle=white,
  colbacktitle=black!60,
  fontupper=\footnotesize,
  boxrule=0.5pt,
  arc=1mm,
  left=3mm,right=3mm,top=3mm,bottom=3mm,
  toptitle=1mm,bottomtitle=1mm,
}

\newcommand{\user}[1]{%
  \vspace{0.4em}
  \noindent
  \begin{tcolorbox}[
    width=\linewidth,
    colback=black!5,
    colframe=black!5,
    boxrule=0pt,
    arc=1mm,
    left=2mm, right=2mm, top=2mm, bottom=2mm,
    standard jigsaw
  ]
  \textbf{User:} #1
  \end{tcolorbox}
  \vspace{0.1em}
}

\newcommand{\model}[1]{%
  \vspace{0.2em}
  \noindent
  \begin{tcolorbox}[
    width=\linewidth,
    colback=white,
    colframe=white,
    boxrule=0pt,
    left=2mm, right=2mm, top=0mm, bottom=0mm,
    sharp corners
  ]
  \textbf{Model:} #1
  \end{tcolorbox}
  \vspace{0.4em}
}

\newcommand{\zerov}[1]{\textcolor{gray!80}{#1}}
\newcommand{\best}[1]{\textcolor{flare4}{#1}}
\newcommand{\res}[3]{#1~\scriptsize{(#2/#3)}}